%% file: forest.tex
\documentclass[11pt]{article}
\usepackage[margin=1.2in]{geometry}

\usepackage[utf8]{inputenc} % allow utf-8 input
\usepackage[T1]{fontenc}    % use 8-bit T1 fonts
\usepackage{hyperref}       % hyperlinks
\usepackage{url}            % simple URL typesetting
\usepackage{booktabs}       % professional-quality tables
\usepackage{amsfonts}       % blackboard math symbols
\usepackage{nicefrac}       % compact symbols for 1/2, etc.
\usepackage{microtype}      % microtypography
\usepackage{multirow}

\usepackage{amssymb,amsbsy,amsmath,amscd,amsthm,amsfonts,bbm}
\usepackage{enumerate, xspace}
\usepackage{graphicx}
\newtheorem{theorem}{Theorem}

\newtheorem{lemma}{Lemma}
\newtheorem{remark}{Remark}

\newtheorem{assumption}{Assumption}

\newcommand{\Binom}{\mathrm{Bin}}

\usepackage{tikz}
\usetikzlibrary{matrix,arrows,calc,shapes,backgrounds}
\usetikzlibrary{shapes.callouts,decorations.text} 
\usetikzlibrary{shapes.misc}

\renewcommand{\tilde}{\widetilde}

\newcommand{\Unif}{\mathrm{Uniform}}

\newcommand{\Expect}{\mathbb{E}}

\newcommand{\Prob}{\mathbb{P}}

\newcommand{\eqdistr}{\,{\buildrel \calD \over =}\,}

\newcommand{\Bern}{\mathrm{Bern}}

\newcommand{\Indc}{\mathbbm{1}}
\newcommand{\indc}[1]{\Indc_{\left\{{#1}\right\}}}

\newcommand{\calC}{{\mathcal{C}}}
\newcommand{\calD}{{\mathcal{D}}}
\newcommand{\calE}{{\mathcal{E}}}

\newcommand{\calM}{{\mathcal{M}}}

\newcommand{\calP}{{\mathcal{P}}}

\newcommand{\calS}{{\mathcal{S}}}

\newcommand{\Th}{{^{\rm th}}}

\usepackage{prettyref,xspace}
\newrefformat{eq}{(\ref{#1})}
\newrefformat{thm}{Theorem~\ref{#1}}
\newrefformat{sec}{Section~\ref{#1}}
\newrefformat{algo}{Algorithm~\ref{#1}}
\newrefformat{fig}{Fig.~\ref{#1}}
\newrefformat{tab}{Table~\ref{#1}}
\newrefformat{rmk}{Remark~\ref{#1}}
\newrefformat{def}{Definition~\ref{#1}}
\newrefformat{cor}{Corollary~\ref{#1}}
\newrefformat{lmm}{Lemma~\ref{#1}}
\newrefformat{prop}{Proposition~\ref{#1}}
\newrefformat{app}{Appendix~\ref{#1}}
\newrefformat{ex}{Example~\ref{#1}}

\DeclareMathOperator*{\argmax}{arg\,max}

\newcommand{\lk}{\lceil\log_2 k_n\rceil}

\usepackage{float}
\usepackage{caption}
\usepackage{subcaption}

\usepackage{xr}
\def\bt{\mathbf{t}}
\def\bx{\mathbf{x}}
\def\bX{\mathbf{X}}

\def\bw{\mathbf{w}}
\def\bm{\mathbf{m}}
\def\btau{\boldsymbol{\tau}}
\def\bbeta{\boldsymbol{\beta}}

\usepackage{mathtools}

\author{
  Jason M.~Klusowski \\
  Department of Statistics\\
  Rutgers University, New Brunswick\\
  Piscataway, NJ, USA, 8019  \\
  \texttt{jason.klusowski@rutgers.edu} \\
}

\title{Sharp analysis of a simple model for random forests}

\begin{document}

\maketitle

\begin{abstract}
\input{abstract}
\end{abstract}

\input{main}

%\subsubsection*{Acknowledgments}
%This work was completed while the author was a visiting graduate student at The Wharton School Department of Statistics. He is grateful to Matthew Olson for suggesting relevant literature to review and Edgar Dobriban for helpful discussions.

\bibliographystyle{plain}
\bibliography{forest}
%\appendix
%\setcounter{equation}{0}
%\setcounter{lemma}{0}

%\renewcommand{\theequation}{\Alph{section}.\arabic{equation}}
%\renewcommand{\thelemma}{\Alph{section}.\arabic{lemma}}
%\renewcommand{\theremark}{\Alph{section}.\arabic{remark}}

%\input{supplement}

\end{document}

%% file: abstract.tex
Random forests have become an important tool for improving accuracy in regression and classification problems since their inception by Leo Breiman in 2001. In this paper, we revisit a historically important random forest model originally proposed by Breiman in 2004 and later studied by G\'erard Biau in 2012, where a feature is selected at random and the splits occurs at the midpoint of the node along the chosen feature. If the regression function is Lipschitz and depends only on a small subset of $ S $ out of $ d $ features, we show that, given access to $ n $ observations and properly tuned split probabilities, the mean-squared prediction error is $ O((n(\log n)^{(S-1)/2})^{-\frac{1}{S\log2+1}}) $. This positively answers an outstanding question of Biau about whether the rate of convergence for this random forest model could be improved. Furthermore, by a refined analysis of the approximation and estimation errors for linear models, we show that this rate cannot be improved in general. Finally, we generalize our analysis and improve extant prediction error bounds for another random forest model in which each tree is constructed from subsampled data and the splits are performed at the empirical median along a chosen feature.

%% file: main.tex
\section{Introduction}
Random forests are ubiquitous among ensemble averaging algorithms because of their ability to reduce overfitting, handle high-dimensional sparse settings, and efficient implementation. 
%As a method that grows many base tree predictors and then combines them, 
%They are related to kernel regression \cite{breiman2000, geurts2006, scornet2016}, adaptive nearest neighbors \cite{lin2006}, and AdaBoost \cite{wyner2017}. 
Due to these attractive features, they have been widely adopted and applied to various prediction and classification problems, such as those encountered in bioinformatics and computer vision.

%Random forests are ubiquitous among ensemble averaging algorithms because of their ability to reduce overfitting and their efficient implementation. They are related to kernel regression \cite{breiman2000, geurts2006, scornet2016}, adaptive nearest neighbors \cite{lin2006}, and AdaBoost \cite{wyner2017}. These connections may explain the success of random forests in various prediction and classification problems, such as those encountered in bioinformatics and computer vision.

One of the most widely used random forests is Breiman's CART algorithm \cite{breiman2001}, which was inspired by the random subspace method of \cite{ho1995}, spatial feature selection of \cite{amit1997}, and random decision method of \cite{dietterich2000}. To this date, researchers have spent a great deal of effort in understanding theoretical properties of various streamlined versions of Breiman's original algorithm \cite{genuer2010, genuer2012, arlot2014, scornet2015, biau2008, denil2014, wager2015, mentch2016, geurts2006}. See \cite{biau2016} for a comprehensive overview of current theoretical and practical understanding. The present paper is an effort to add to this body of work.

We assume the training data is $ \calD_n = \{(\bX_1, Y_1), \dots, (\bX_n, Y_n) \} $, where $(\bX_i , Y_i ) $, $1\leq i \leq n $ are i.i.d. with common joint distribution $\mathbb{P}_{\bX, Y} $. Here, $\bX_i \in [0, 1]^d $ is the feature or covariate and $Y_i \in \mathbb{R} $ is a continuous response variable. 
%A generic pair of variables will be denoted as $ (\bX, Y) $ with joint distribution $\mathbb{P}_{\bX, Y} $. 
The $ j^\text{th} $ feature of $ \bX $ will be denoted by $ \bX^{(j)} $. We make the following assumptions on the statistical regression model.
\begin{assumption} \label{ass:data}
The response variable can be written as $Y_i = f(\bX_i) + \varepsilon_i $, for  $ i = 1, \dots, n $ where $ f(\bx) = \Expect[Y \mid \bX = \bx] $ is an unknown regression function and $\{ \varepsilon_i \}_{1 \leq i\leq n} $ are i.i.d. errors. Furthermore, $ \Expect[Y^2] < \infty $ and $ \text{VAR}(Y \mid \bX) \equiv \sigma^2 $, for some positive constant $\sigma^2$, and $ \bX $ is uniformly distributed on $ [0, 1]^d $.
\end{assumption}

\begin{assumption} \label{ass:function}
The regression function $ f(\cdot) $ is bounded in magnitude by a positive constant $ B $ and has bounded first-order partial derivates, i.e., $ \|\partial f_j\|_{\infty} \coloneqq \sup_{\bx\in[0,1]^d}|\partial_j f(\bx)| < \infty $ for $ j = 1, 2, \dots, d $. The largest infinity norm of the partial derivatives is denoted by $ L = \max_j  \|\partial f_j\|_{\infty} $.
\end{assumption}
%The conditional average of $ Y $ given $ \bX $ is optimal for prediction if one uses squared error loss $ L(Y, Y') = (Y-Y')^2 $ since it minimizes the conditional risk $ \Expect[(Y-\widetilde f(\bX))^2 \mid \bX] = \int |y-\widetilde f(\bX)|^2 \mathbb{P}_{Y \mid \bX}(dy) $. Therefore, our goal throughout this paper is to predict the conditional mean response of $ Y $ at a new observation $ \bX = \bx $ from $ \calD_n $.
 The efficacy of a predictor $ \widehat Y(\bx) = \widehat Y(\bx;\calD_n) $ of $ f(\cdot) $ will be measured in terms of its \emph{mean squared prediction error}, $ \Expect[(\widehat Y(\bX)-f(\bX))^2] $, where the expectation is with respect to the new input $ \bX $ and the training data $ \calD_n $. Throughout this paper, $ \lambda $ is the Lebesgue measure.
%For example, in a sparse single-index model, $ f(x) = g(\bbeta \cdot x) $, where $ g $ is a nonparametric link function and $ \bbeta $ is a $ d $-dimensional, $ S $ sparse vector. 
%Per the success of many modern methods in machine learning, the sparsity assumption corresponds to the fact that a large class of high-dimensional functions admit or are approximated well by sparse representations of simple model forms, e.g., wavelets, sparse linear combinations of nonlinear ridge compositions, or neural networks with sparse weights. 

As mentioned earlier, many scholars have proposed and studied idealized versions of Breiman's original algorithm \cite{breiman2001}, largely with the intent of reducing the complexity of their theoretical analysis. Unlike Breiman's CART algorithm, these stylized versions are typically analyzed under the assumption that the probabilistic mechanism $ \Theta $ that governs the construction of each tree \emph{does not depend} on the training sample $ \calD_n $ (i.e., the splits are not data dependent). Such models are referred to as \emph{purely random forests} \cite{genuer2012}.
 On the other hand, recent works have proved properties like asymptotic normality \cite{mentch2016, wager2014} or consistency \cite{scornet2015, scornet2016asymptotics, denil2014}, where the data may be bootstrapped or the splits determined by optimizing some empirical objective.
 %\footnote{For example, at each step, CART chooses the split that optimizes a conditional sum of squares criterion.} 
However, these results are asymptotic in nature, and it is difficult to determine the quality of convergence as a function of the parameters of the random forest (e.g., sample size, dimension, and depth to which the individual trees are grown). 

In this paper, we focus on another historically significant model that was proposed by Breiman in a technical report \cite{breiman2004}. Here, importantly, the individual trees are grown \emph{independently of the training sample} $\calD_n$ (although subsequent work allows the trees to depend on a second sample $ \calD'_n $, independent of $ \calD_n $).
% \emph{independently of $\bX$ and the training sample $\calD_n$}
Despite its simplicity, this random forest model captures a few of the attractive features of Breiman's original algorithm \cite{breiman2004}, i.e., variance reduction by randomization, and adaptive feature selection. This model also allows us to provide a non-asymptotic prediction error bound that reveals the dependence on the parameters of the forest.

%The tree construction can be described by repeating the following procedure until a desired depth has been achieved.
%\begin{enumerate}[(i)]
%\item At each node, select a feature of $\bX=(\bX^{(1)}, \dots, \bX^{(d)})$ at random, with the $j^{\text{th}}$ feature having a %probability $p_{j}$ of being selected, where $ \sum_{j=1}^dp_{j} = 1 $.
%\item If the variable is strong, split at the midpoint of the selected variable at the node.
%\item If the variable is weak, split at a random point along its values in the node.
%\end{enumerate}
Later, in an influential paper, \cite{biau2012} considered the same model and rigorously established some informal, heuristic-based claims made by Breiman. Both works of Breiman and Biau will serve as the basis for this article, whose primary purpose is to strengthen the analysis of this model and offer a full picture of its fundamental limits. Borrowing the terminology of \cite{scornet2016}, we shall refer to this model henceforth as a \emph{centered random forest}. In the forthcoming discussion, $ \log $ is the natural logarithm.

\paragraph{New contributions.} 
%As with Biau, we focus on a variant of the above model, where we do not distinguish between strong or weak features and always %split at the midpoint of the selected variable at the node.
%\begin{enumerate}[(i)]
%\item At each node, select a feature of $\bX=(\bX^{(1)}, \dots, \bX^{(d)})$ at random, with the $j^{\text{th}}$ feature having a probability $p_{j}$ of being selected, where $ \sum_{j=1}^dp_{j} = 1 $.
%\item Split at the midpoint of the selected variable at the node.
%\end{enumerate}xf
To avoid the curse of dimensionality—which plagues high-dimensional regression models—and the associated undesirable consequences (e.g., overfitting and large sample requirements), it is typically assumed that $ f(\cdot) $ is sparse in the sense that it only depends on a small subset $ \calS $ of the $ d $ features, where $ S \coloneqq |\calS| \ll d $. In other words, $ f(\cdot) $ is almost surely equal to its restriction to the subspace of its \emph{strong} features in $ \calS $. Conversely, the output of $ f(\cdot) $ does not dependent on \emph{weak} features that belong to $ \calS^c $. Of course, the set $ \calS $ is not known a priori and must be learned from the data. Within this framework, \cite[Corollary 6]{biau2012} showed that with properly tuned probabilities that each feature is split, the mean squared prediction error is
\begin{equation} \label{eq:biaurate}
O(n^{-\frac{1}{S(4/3)\log 2+1}}).
\end{equation}
A surprising aspect of \prettyref{eq:biaurate} is that the exponent is independent of the ambient dimension $ d $, which might partially explain why random forests perform well in high-dimensional sparse settings. Biau also raised the question \cite[Remark 7]{biau2012} as to whether this rate could be improved. We will answer this in the affirmative and show that the error \prettyref{eq:biaurate} can indeed be improved to
\begin{equation} \label{eq:rate}
O((n\log^{(S-1)/2}n)^{-\alpha_S}),
\end{equation}
where
\begin{equation*}
\alpha_S \coloneqq \frac{2\log_2(1-S^{-1}/2)}{2\log_2(1-S^{-1}/2)-1} = \frac{1}{S\log 2 + 1}(1+\Delta_S),
\end{equation*}
and $ \Delta_S $ is some positive quantity that decreases to zero as $ S $ approaches infinity.
In particular,
\begin{enumerate}[(a)]
\item We improve the rate in the exponent from $ \frac{1}{S(4/3)\log 2+1} $ to $ \frac{1}{S\log 2 + 1} $ and, due to the presence of the logarithmic term in \prettyref{eq:rate}, improve the convergence by a factor of $ O((\log n)^{-\frac{1}{2\log 2}}) $. Note that the rate \prettyref{eq:rate} is \emph{not} minimax optimal for the class of Lipschitz regression functions in $ S $ dimensions, unless $ S = 1 $.
\item We generalize our proof techniques and use them to improve the convergence rates of other random forest models. In particular, for \emph{median random forests} \cite{duroux2018impact}, we improve the rate from $ O(n^{-\frac{\log_2(1-3d^{-1}/4)}{\log_2(1-3d^{-1}/4)-1}}) $ to $ O(n^{-\frac{2\log_2(1-d^{-1}/2)}{2\log_2(1-d^{-1}/2)-1}}) $.
% In particular, centered random forests have variance of order $ O((\sqrt{\log n})^{-(S-1)}) $, even when each terminal node contains, on average, only a single observation.
\item We show that the rate \prettyref{eq:rate} is not generally improvable for centered random forests. To accomplish this, we show that the approximation error is tight for all linear models with nonzero parameter vector. We also characterize the estimation error, which is, surprisingly, nearly the smallest among \emph{all} purely random forests with nonadaptive splitting schemes.
% The lower bounds, which are minimax suboptimal, also stress the importance of having a data-dependent split protocol and feature selection criterion and the crucial role they play in determining the good performance of forests implemented in practice.
%\item We show that if the regression function is square-integrable (e.g., it need not be continuous or even bounded), then the random forest predictor is pointwise consistent almost everywhere.
\end{enumerate}
Additional comparisons between our work and \cite{biau2012} and \cite{duroux2018impact} are provided in \prettyref{tab:rates}. The improvements in (a) and (b) stem from a novel analysis of the estimation and approximation errors of the random forest.
% and of the correlation between trees. We also believe our new techniques are generalizable and can be used to improve existing results for other random forest models, i.e., improve the approximation error/approximation error of \emph{median forests} \cite{duroux2018impact} and the subpar mean squared prediction error $ O(n^{-\frac{\log(1-0.75/d)}{\log(1-0.75/d)-\log 2}}) $ \cite[Theorem 3.1]{duroux2018impact}, \cite[Theorem 4]{wager2015}. 
%to $ n^{-\alpha_S} $ for $ S = d $.

\paragraph{Related results.} We now mention a few related results. \cite{scornet2016} slightly altered the definition of random forests so that they could be rewritten as kernel methods. \cite[Theorem 1]{scornet2016} showed that \emph{centered kernel random forests}, where the trees are grown according to the same selection and splitting procedure as centered random forests, have mean squared prediction error $ O(n^{-\frac{1}{d\log 2+3}}\log^2 n) $. In addition to the computational advantages of centered random forests when $ n $ and $ d $ are moderately sized, note that \prettyref{eq:rate} is strictly better. The improved rate \prettyref{eq:rate} is obtained by growing the trees to a shallower depth than the depth used by Scornet, and this may explain why the author found centered kernel random forests to empirically outperform centered random forests for certain regression models \cite[Model 1, Figure 5]{scornet2016}.
%It would be interesting to rerun his experiments with our new depth level.

Other results have been established for function classes with additional smoothness assumptions. For example, a multivariate function on $ [0,1]^d $ is of class $ \calC_k([0, 1]^d) $ if all its $ k\Th $ order partial derivatives exist and are bounded on $ [0, 1]^d $. Then, for regression functions in $ \calC_2([0, 1]^d) $, \cite[p. 21]{arlot2014} obtained a similar rate of $ O(n^{-\alpha_S}) $ for $ S = d \geq 4 $ under the so-called \emph{balanced purely random forest} model, where all nodes are split at each stage (in contrast to single splits with centered random forests). However, in addition to requiring that the regression function is of class $ \calC_2([0,1]^d) $ (instead of just Lipschitz), it is unclear whether these random forest models can be modified to adapt to sparsity.
%(i.e., where only a subset of the variables have an effect on the output). 
%It would be interesting to see if our new techniques could be used to remove the $ \calC_2([0,1]^d) $ condition so that the same rate also holds for $ \calC_1([0,1]^d) $.

Finally, there are online versions of random forests, albeit defined somewhat differently than centered random forests, which perform better. Recently, \cite{mourtada2018} have shown that a type of online forest known as \emph{Mondrian forests} achieve minimax optimal rates when $ f(\cdot) $ belongs to $ \calC_1([0,1]^d) $ or $ \calC_2([0,1]^d) $, i.e., $ \Theta(n^{-\frac{2}{d+2}}) $ or $ \Theta(n^{-\frac{4}{d+4}}) $, respectively \cite[Example 6.5]{yang1999}.

\paragraph{Organization.} This paper is organized as follows. We briefly review basic terminology of decision tree ensembles and define centered and median random forests in \prettyref{sec:centered}. In \prettyref{sec:main}, we present our main results, which are derived from an analysis of the approximation and estimation errors of the random forest. In \prettyref{sec:tight}, we show that the approximation and estimation error bounds derived in \prettyref{sec:main} cannot be generally improved. Proofs of all supporting lemmas are given in \prettyref{app:appendix}.
%\prettyref{sec:improve} considers a toy example to illustrate how better performance can be attained if data-dependent splits are used to grow each tree. 

\section{Random forests} \label{sec:centered}

In general terms, a random forest is a predictor that is built from an ensemble of randomized base regression trees $\{\widehat Y(\bx; \Theta_m,\calD_n) \}_{1 \leq m \leq M}$. The sequence $ \{\Theta_m\}_{1 \leq m\leq M} $ consists of i.i.d. realizations of a random variable $\Theta$, which governs the probabilistic mechanism that builds each tree. These individual random trees are aggregated to form the final output
\begin{equation*}
\widehat Y_M(\bX; \Theta_1, \dots, \Theta_M, \calD_n) \coloneqq \frac{1}{M}\sum_{m=1}^M \widehat Y(\bX; \Theta_m, \calD_n).
\end{equation*}
When $ M $ is sufficiently large, Theorem 3.3 from \cite{scornet2016asymptotics} justifies using
\begin{equation*}
\widehat Y(\bX) = \widehat Y(\bX, \calD_n) \coloneqq \Expect_{\Theta}[\widehat Y(\bX; \Theta, \calD_n)],
\end{equation*}
in lieu of $ \widehat Y(\bX; \Theta_1, \dots, \Theta_M, \calD_n) $, where $\Expect_{\Theta}$ denotes expectation with respect to $ \Theta $, conditionally on $\bX$ and $\calD_n$. We  henceforth work with this asymptotic random forest.

The randomized base regression tree $ \widehat Y(\bX; \Theta, \calD_n) $ is a local weighted average of all $ Y_i $ for which the corresponding $ \bX_i $ falls into the same node of the random partition as $ \bX $. For concreteness, let $ \bt = \bt(\bX, \Theta, \calD_n) $ be the node of the random partition
containing $ \bX $ and define the individual tree predictor via
\begin{equation*}
\widehat Y(\bX; \Theta, \calD_n)\coloneqq\frac{\sum_{i=1}^n Y_i \indc{\bX_i \in \bt}}{\sum_{i=1}^n \indc{\bX_i \in \bt}}\,\Indc_{\calE},
\end{equation*}
where $ \calE $ is the event that $ \sum_{i=1}^n \indc{\bX_i \in \bt} $ is nonzero.
We then take the expectation of these individual predictors with respect to the randomizing variable $ \Theta $ yielding
\begin{equation*}
\widehat Y(\bX) = \sum_{i=1}^n \Expect_{\Theta} [W_{i}]Y_i,
\end{equation*}
where 
\begin{equation*} 
W_{i} = W_i(\bt) \coloneqq\frac{\indc{\bX_i \in \bt}}{N(\bt)}\,\Indc_{\calE}
\end{equation*}
are the weights corresponding to each observed output and 
\begin{equation*} 
N(\bt)\coloneqq\sum_{i=1}^n \indc{\bX_i\in \bt}
\end{equation*}
is the total number of observations that fall into the same box of the random partition as $\bX$. The node $ \bt $ is a Cartesian product and thus can be decomposed into the product of its sides $ \prod_{j=1}^d [a_j, b_j]$, where $ a_{j} = a_j(\bX, \Theta, \calD_n) $ and $ b_{j} = b_j(\bX, \Theta, \calD_n) $ are its left and right endpoints, respectively, along the $ j^{\text{th}} $ axis.

%We will assume for the sake of exposition that the feature-sampling probabilities satisfy the constraints $ p_{j} = 1/S $ for $ j \in \calS $ and $ p_{j} = 0 $ otherwise, despite that fact that we do no know the set $ \calS $ a priori. 

Let us now formally define how each base tree $ \widehat Y(\bx; \Theta_m,\calD_n) $ of a centered random forest and median random forest are constructed. We first describe the centered random forest from \cite{breiman2004} and \cite{biau2012}.
%Every node of the tree has a corresponding box (i.e., $ d $-dimensional hyperrectangle) and at each stage of the construction of the tree, the collection of terminal nodes of the tree forms a partition of $ [0,1]^d $. 
%In fact, we will see that the boxes are dyadic cubes.

\paragraph{\bf{Centered random forest.}}
\begin{enumerate}[(i)]
\item Initialize with $ [0,1]^d $ as the root.
\item At each node, select one feature $ j $ in $ \{1, 2, \dots, d \} $ with probability $(p_j)_{1 \leq j\leq d}$, where $ \sum_{j=1}^dp_{j} = 1 $.
\item Split the node at the midpoint of the interval along the direction of the selected feature.
\item Repeat steps (ii) and (iii) for the two daughter nodes until each node has been split exactly $\lk$ times.
%\footnote{Its value will be determined by optimizing the tradeoff between the variance and approximation error of $ \widehat Y(\bX) $}
\end{enumerate}

\begin{remark}
Let us briefly mention that this model is similar in spirit to a recent random forest model proposed by \cite{basu2018iterative}, coined \emph{iterative random forests}. Iterative random forests explicitly learn feature sampling probabilities, and so the results from the present paper could be useful for studying a simplified variant of the model.
\end{remark}

The split probabilities $ (p_j)_{1 \leq j\leq d} $ determine how frequently a particular direction is split. By tuning these probabilities to be large for strong directions in $ \calS $ and small otherwise, one can show convergence rates that do not degrade severely with the ambient dimension $ d $. In \prettyref{sec:data}, we will consider data-driven choices of $ (p_j)_{1 \leq j\leq d} $ with the aide of a second sample $ \calD'_n $, independent of $ \calD_n $. In this case, the probabilities are \emph{data-dependent}, i.e., $ p_j = p_j(\calD'_n) $, and therefore our forthcoming prediction error bounds are written conditional on them.

%\begin{enumerate}
%\item \label{fact:fact1} Since by construction, $ \sum_{j=1}^d K_{j}=\lk $, we have that, conditionally on $ \bX $, $ (K_{n1}, \dots, K_{nd}) $ follows a multinomial distribution with $\lk$ trials and event probabilities $ (p_j)_{1 \leq j\leq d}$. Note that \cite{biau2012} only needs that the marginals $ K_{j} $ are binomially distributed $ \Binom(p_{j}, \lk) $, but we will need to work with their joint distribution.

%\item \label{fact:fact2} Since by construction, $ N(\bt)= \sum_{i=1}^n \indc{\bX_i\in \bt} $ and $ \lambda (\bt)=2^{-\lk} $, we have that conditionally on $\bX$ and $\Theta$, $N(\bt)$ is binomial with $n$ trials and success probability $2^{-\lk}$.
%\end{enumerate}
The next random forest model we study is similar to centered random forests, though there are two important differences. First, each tree is constructed from subsampled data and, second, the splits are performed at the empirical median in an interval along a randomly chosen feature—thus making the splits \emph{data-dependent}. As we will see, if the split probability sequence $ (p_j)_{1 \leq j \leq d} $ is uniform over all $ d $ features, these two random forest models have nearly identical convergence rates.

\paragraph{\bf{Median random forest.}}

\begin{enumerate}[(i)]
\item Select, uniformly without replacement, $n_0<n$ data points among $\mathcal{D}_n$. Only these $n_0$ observations will be used in the tree construction. 
\item Initialize with $ [0,1]^d $ as the root.
\item At each node, select uniformly at random one feature $ j $ among $ \{1, 2,\dots, d\} $. 
\item Split the node at the empirical median of the $\bX^{(j)}_i$ in the interval along the selected feature.  
\item Repeat steps (iii) and (iv) for the two daughter nodes until each node has been split exactly $\lk$ times.
\end{enumerate}

\begin{remark}
Since $ \bX^{(j)} $ is uniformly distributed on $ [0, 1] $, it has a binary expansion
\begin{equation*}
\bX^{(j)} \eqdistr \sum_{k \geq 1} B_k 2^{-k},
\end{equation*}
where $ \{B_k\}_{k=1}^{\infty} $ are i.i.d. $ \Bern(1/2) $. Thus, for the centered random forest model, if $ K_j = K_j(\bX, \Theta) $ is the number of times the nodes are split along the $ j^{\text{th}} $ feature, each endpoint of $ [a_j, b_j] $ is a randomly stopped binary expansion of $ \bX^{(j)} $, viz.,
\begin{equation} \label{eq:lend}
a_{j} \eqdistr \sum_{k =1}^{K_{j}} B_k 2^{-k},
\end{equation}
and
\begin{equation} \label{eq:rend}
b_{j} \eqdistr 2^{-K_{j}} + \sum_{k =1}^{K_{j}} B_k 2^{-k}.
\end{equation}

%A routine calculation reveals that these endpoints concentrate around
%\begin{align*}
%\frac{1}{2}(1-(1-p_{j}/2)^{\lk}) \qquad \mbox{and} \qquad \frac{1}{2}(1+(1-p_{j}/2)^{\lk}),
%\end{align*}
%respectively. 
%As we have said before, each randomized base regression tree is a local weighted average of all $ Y_i $ for which the corresponding $ \bX_i $ falls into the same node of the random partition as $ \bX $, but in light of \prettyref{eq:lend} and \prettyref{eq:rend}, we may also view it as a local weighted average of all $ Y_i $ for which the first $ K_{j} $ entries of the binary strings $ (B_{j}(\bX_i))_{j=1}^{\infty} $ and $ (B_{j}(\bX))_{j=1}^{\infty} $ are equal. 

The representations \prettyref{eq:lend} and \prettyref{eq:rend} will also prove to be useful when we derive converse statements for this random forest model.
\end{remark}

Armed with these concepts and notation, we are now ready to present our main results.
\section{Main results} \label{sec:main}

We begin our analysis with the standard approximation/estimation error decomposition of the mean squared prediction error:
\begin{equation}
\label{eq:bv}
\mathbb{E}[(\widehat Y(\bX)- f(\bX))^2]= \underbrace{\mathbb{E}[(\overline Y(\bX) - f(\bX))^2]}_{\mbox{approximation error}} + \underbrace{\mathbb{E}[(\widehat Y(\bX)-\overline Y(\bX))^2]}_{\mbox{estimation error}},
\end{equation}
where $ \overline Y(\bX)  \coloneqq \mathbb{E}[\widehat Y(\bX)\mid \bX_1, \dots, \bX_n, \bX] $. As is generally true with nonadaptive random forests, the estimation error is typically of order $ \sigma^2 k_n/n $. What does vary with the specific random forest model, however, is the approximation error. Below we give a general upper bound on the approximation error that is valid for any random forest model.

\begin{theorem} \label{thm:approximation error}
%Suppose $ f(\cdot) $ is $ L $-Lipschitz and has $ \mathbb{L}^{\infty} $ norm at most $ B $. 
%Suppose the joint distribution of the tree $ \Theta $ and input training data $ \bX_1, \dots, \bX_n $ is symmetric in permutations of the training data index, i.e.,
%$$
%(\bX_1, \dots, \bX_n, \Theta) \eqdistr (\bX_{\tau(1)}, \dots, \bX_{\tau(n)}, \Theta), \quad \text{for all permutations}\; \tau: [n] \mapsto [n].
%$$
For any random forest model, under \prettyref{ass:function},
\begin{equation} \label{eq:approx}
\mathbb{E}[(\overline Y(\bX) - f(\bX))^2] \leq S\sum_{j=1}^d \|\partial f_j\|^2_{\infty}\Expect[(\Expect_{\Theta}[b_{j}-a_{j}])^2] + B^2\mathbb{P}(\calE^c).
\end{equation}
\end{theorem}

\begin{proof}
We first decompose the approximation error $\mathbb{E}[(\overline Y(\bX) - f(\bX))^2] $ as follows:
\begin{align}
& \Expect\Big[\Big(\sum_{i=1}^n\Expect_{\Theta}[W_{i}(f(\bX_i) - f(\bX))] - \indc{\calE^c}f(\bX) \Big)^2\Big] \nonumber \\
& = \Expect\Big[\Big(\sum_{i=1}^n\Expect_{\Theta}[W_{i}(f(\bX_i) - f(\bX))]\Big)^2\Big] + \Expect[\indc{\calE^c}|f(\bX)|^2] \label{eq:iden1} \\
& \leq \Expect\Big[\Big(\sum_{i=1}^n\Expect_{\Theta}[W_{i}(f(\bX_i) - f(\bX))]\Big)^2\Big] + B^2\Prob(\calE^c). \label{eq:last}
\end{align}
Next, by \prettyref{ass:function}, we have that $ |f(\bX_i) - f(\bX)| \leq \sum_{j=1}^d \|\partial_j f\|_{\infty}|\bX^{(j)}_i -\bX^{(j)}|  $, and thus, $ W_{i}|f(\bX_i) - f(\bX)| \leq W_{i}\sum_{j=1}^d \|\partial_j f\|_{\infty}(b_{j}-a_{j})  $. This shows that 
\begin{align*}
\sum_{i=1}^nW_{i}|f(\bX_i) - f(\bX)| & \leq \sum_{i=1}^nW_{i}\sum_{j=1}^d \|\partial_j f\|_{\infty}(b_{j}-a_{j}) \\
& \leq \sum_{j=1}^d \|\partial_j f\|_{\infty}(b_{j}-a_{j}).
\end{align*}
Taking expectations with respect to $ \Theta $ of both sides of this inequality, we may bound the first term in \prettyref{eq:last} by
$$
\Expect\Big[\Big(\sum_{j=1}^d \|\partial_j f\|_{\infty}\Expect_{\Theta}[b_{j}-a_{j}]\Big)^2\Big].
$$
The Cauchy-Schwarz inequality then yields
$$
S\sum_{j=1}^d \|\partial_j f\|^2_{\infty}\Expect[(\Expect_{\Theta}[b_{j}-a_{j}])^2].
$$
\end{proof}
Despite its simple proof, \prettyref{thm:approximation error} leads to nontrivial improvements over past work. It is now easy to isolate precisely where our improvements manifest. In standard analysis of random forest models, the quantity $ \Expect_{\Theta}[(b_{j}-a_{j})^2] $ is typically analyzed directly, where the $ \Theta $-averaging occurs on the outside of the square. On the other hand, the bound \prettyref{eq:approx} allows the $ \Theta$-averaging to occur \emph{inside} the square, and thus by Jensen's inequality, it represents a uniform improvement, i.e.,
$$
(\Expect_{\Theta}[b_{j}-a_{j}])^2 \leq \Expect_{\Theta}[(b_{j}-a_{j})^2].
$$
Both \cite{biau2012} and \cite{duroux2018impact} bound the approximation error by $ O(k^{\log_2(1-3d^{-1}/4)}_n) = O(k^{-\frac{1}{d(4/3)\log 2}}_n) $. We will use \prettyref{eq:approx} to improve this bound to $ O(k^{2\log_2(1-d^{-1}/2)}_n) = O(k^{-\frac{1}{d\log 2}}_n) $. Note that this bound is the same (up to a constant factor) as \cite[Corollary 9]{arlot2014} when $ d \geq 4 $, though the authors analyze the balanced purely random forest model and make a stronger assumption that $ f(\cdot) $ has bounded second-order partial derivatives. 

%Each of these terms will be controlled in \prettyref{thm:variance} and \prettyref{thm:approximation error}. The next result is proved by combining the bounds in \prettyref{thm:variance} and \prettyref{thm:approximation error} with the variance/approximation error decomposition in \prettyref{eq:bv}.
%We let $ \calS = \{ j : p_{j} \neq 0 \} $.
%In accordance with the discussion in \prettyref{sec:centered}, we assume throughout that $ p_{j} = (1/S)(1+{\xi}_{j}) $ for $ j \in \calS $ and $ p_{j} = {\xi}_{j} $ otherwise, where $ \{{\xi}_{j} \} $ is a sequence that tends to zero as $ n $ tends to infinity. Furthermore, let $ p_n = (1/S)(1+\xi_n) $, where $ \xi_n = \min_{j\in\calS} \xi_{j} $.
\subsection{Centered random forests}
In this subsection, we derive bounds on the mean squared prediction error of a centered random forest in terms of $ k_n $ and the probability sequence $ (p_j)_{1 \leq j\leq d} $. As a consequence, we also obtain rates of convergence.

\begin{theorem}[Centered random forests] \label{thm:centerrisk}
Let $ \calP \coloneqq \{ j : p_j \neq 0 \} $ and $ d_0 \coloneqq \#\calP $. Under \prettyref{ass:data} and \prettyref{ass:function} and conditional on $ (p_j)_{1 \leq j\leq d} $,
\begin{align*}
\mathbb{E}[(\widehat Y(\bX)- f(\bX))^2]& \leq S\sum_{j=1}^d\|\partial_j f \|^2_{\infty}k^{2\log_2(1-p_{j}/2)}_n \\ & \qquad + \frac{12\sigma^2k_n}{n}
\frac{8^{d_0}}{\sqrt{\prod_{j\in\calP}p_j \times \log^{d_0-1}_2(k_n)}}
 +  B^2e^{-n/(2k_n)}. %\frac{(8S)^{S-1}}{(1+\xi_n)^{S-1}\sqrt{\log^{S-1}_2 k_n}}.
\end{align*}
Consequently, if $ p \coloneqq \min_j p_j $, $ \alpha \coloneqq \frac{2\log_2(1-p/2)}{2\log_2(1-p/2)-1}  $, and $ k_n \asymp (n(\log^{d-1}_2 n)^{1/2})^{1-\alpha} $, then, conditional on $ p $, there exists a constant $ C > 0 $, depending only on B, $ S $, $ d $, $ L $, and $ \sigma^2 $ such that 
$$
\mathbb{E}[(\widehat Y(\bX)- f(\bX))^2] \leq C(n(\log^{d-1}_2 n)^{1/2})^{-\alpha}.
$$
\end{theorem}

\begin{proof}

%\paragraph{Approximation error.}
First, \cite[Section 5.3, p. 1089]{biau2012} shows that $ \mathbb{P}(\calE^c) \leq e^{-n/(2k_n)} $. Next, let $ K_j = K_j(\bX, \Theta) $ be the number of times the nodes are split along the $ j^{\text{th}} $ feature and note that $ K_j $ is conditionally distributed $ \Binom(p_{j}, \lk) $ given $ \bX $. Then, conditional on $(p_j)_{1 \leq j\leq d}$,
\begin{align*}
\Expect_{\Theta}[b_{j}-a_{j}] & = \Expect_{\Theta}[2^{-K_{j}}] = \Expect_{K\sim\Binom(p_{j},\lk)}[2^{-K}] \\
& = (1-p_{j}/2)^{\lk}
 \leq k_n^{\log_2(1-p_{j}/2)}.
\end{align*}
Thus, by \prettyref{thm:approximation error}, the approximation error is bounded by 
\begin{equation} \label{eq:bias}
\mathbb{E}[(\overline Y(\bX) - f(\bX))^2] \leq S\sum_{j=1}^d \|\partial_j f\|_{\infty}k_n^{2\log_2(1-p_j/2)} + B^2 e^{-n/(2k_n)}.
\end{equation}
%\paragraph{Estimation error.} 
Next, we bound the estimation error of the random forest. In particular, we show that,
%\begin{lemma} \label{thm:variance}
conditional on $ (p_j)_{1 \leq j \leq d} $,
\begin{equation} \label{eq:variance}
\mathbb{E}[(\widehat Y(\bX)-\overline Y(\bX))^2] \leq \frac{12\sigma^2k_n}{n}\frac{8^{d_0}}{\sqrt{\prod_{j\in\calP}p_j \times \log^{d_0-1}_2 k_n}}.
\end{equation}
%\end{lemma}
%It is possible that the lower bound \prettyref{eq:varlower} is not tight and can be improved by a square-root in the exponent of the logarithm.

%\begin{proof}[Proof of \prettyref{thm:variance}]
Henceforth, we let $ K'_j $, $ [a_j', b_j'] $, and $ \bt' $ denote the feature selection frequency, terminal node side, and terminal node, respectively, from an independent copy $ \Theta' $ of $ \Theta $. It is shown in \cite[Section 5.2, p. 1085]{biau2012} that
\begin{equation} \label{eq:varmain}
\mathbb{E}[(\widehat Y(\bX)-\overline Y(\bX))^2] \leq \frac{12\sigma^2k^2_n}{n}\Expect_{\Theta, \Theta'}[\lambda(\bt\cap \bt')].
\end{equation}
We can use the representations \prettyref{eq:lend} and \prettyref{eq:rend} to show that for any $ \Theta $ and $ \Theta' $, the sides of the node are nested according to $ [a'_j, b'_j] \subseteq [a_j, b_j] $ if and only if $ K_{j} \geq K'_j $ and hence
\begin{equation} \label{eq:subset}
\lambda([a_j, b_j] \cap [a'_j, b'_j]) = 2^{-\max\{K_{j}, K'_j \}}.
\end{equation}
Using this, we have
\begin{align} \label{eq:capiden}
\lambda(\bt\cap \bt') & = \prod_{j=1}^d \lambda([a_j, b_j]\cap [a'_j, b'_j]) \nonumber \\
& = 2^{-\sum_{j=1}^d \max\{ K_{j}, K'_j\} }
 = 2^{-\lk-\frac{1}{2}\sum_{j=1}^d | K_{j}- K'_j|},
\end{align}
where the equality in \prettyref{eq:capiden} follows from the identity
\begin{align*}
\sum_{j=1}^d \max\{ K_{j}, K'_j\} & = \frac{1}{2}\sum_{j=1}^d K_{j} + \frac{1}{2}\sum_{j=1}^d K'_j + \frac{1}{2}\sum_{j=1}^d | K_{j}- K'_j| \\ 
& = \lk + \frac{1}{2}\sum_{j=1}^d | K_{j}- K'_j|.
\end{align*}
Next, note that conditional on $ \bX $, $ (K_1, \dots, K_d) $ has a multinomial distribution with $ \lk $ trials and event probabilities $ (p_j)_{1 \leq j\leq d} $.
%\footnote{In an earlier draft of this paper, we experimented with a random number of terminal nodes $ k_n \sim \Poisson(\lambda) $, so that $ \{ K_{j} \}_{j \in \calS} $ are independent $ \Poisson(\lambda p_{j}) $. Then the expected value of the product \prettyref{eq:capiden} is simply the product of the expectations, which involves only the marginals $ K_{j} $. Despite the advantages of independence, it can be shown that this random forest has worse mean squared predictive error, even when the approximation error and variance are optimized over $ \lambda $.
%approximation error $ O(2^{-\frac{\lambda p_n}{\log 2}}) $ and variance $ O(n^{-1}\lambda^{-p^{-1}_n/2}2^{\lambda(1+(\log 2)/4)}) $, which leads to a suboptimal rate $ O((n(\sqrt{\log n})^S)^{-\frac{p_n}{\log 2(1+(\log 2)/4)+p_n}}) $.
%}
%Thus, we have the 
%\begin{align}
%\expect{2^{-\sum_{j=1}^d \max\{ K_{j}, K'_j\}}} & = 2^{-\lk}\expect{2^{-\frac{1}{2}\sum_{j=1}^d | %K_{j}- K'_j|}} \label{eq:maxiden} \\
%& \leq k^{-1}_n\expect{2^{-\frac{1}{2}\sum_{j=1}^d | K_{j}- K'_j|}}, \nonumber
%\end{align}
We take the expected value of \prettyref{eq:capiden} and use the bound 
%\begin{equation}
%\Expect[2^{-\frac{1}{2}\sum_{j=1}^d | K_{j}- K'_j|}] \leq \Expect[2^{-\frac{1}{2}\sum_{j=1}^d | K_{j}- K'_j|}]. \label{eq:Kbound}
%\end{equation}
%Next, let $ \{j_1, j_2, \dots, j_S\} $ be an enumeration of $ \calS $. By 
\prettyref{eq:multibound} (whose proof is given in \prettyref{lmm:multi}), yielding
\begin{align}
\Expect_{\Theta, \Theta'}[2^{-\frac{1}{2}\sum_{j=1}^d | K_{j}- K'_j|}] = \Expect_{\Theta, \Theta'}[2^{-\frac{1}{2}\sum_{j\in\calP} | K_{j}- K'_j|}]
%& \leq \frac{8^{S-1}}{\sqrt{(\log^{S-1}_2 k_n)p^{S-1}_{nj_S}\prod_{k=1}^{S-1}p_{j_k}}} \nonumber \\
& \leq \frac{8^{d_0}}{\sqrt{\prod_{j\in\calP}p_j \times \log^{d_0-1}_2 k_n}}. \label{eq:corrbound}
\end{align}
Combining \prettyref{eq:varmain}, \prettyref{eq:capiden}, and \prettyref{eq:corrbound} proves \prettyref{eq:variance}.
\end{proof}

\begin{remark}
In proving the estimation error bound \prettyref{eq:variance}, we depart from the strategy of \cite{biau2012}, which we now briefly outline. Biau's approach consists of applying H\"older's inequality to the expectation of \prettyref{eq:capiden} and resultant expected product, i.e.,
\begin{align*}
\Expect_{\Theta, \Theta'}[2^{-\sum_{j=1}^d \max\{ K_{j}, K'_j\}}]
& \leq k^{-1}_n\prod_{j\in\calP}(\Expect_{\Theta, \Theta'}[2^{-\frac{d}{2}|K_{j}-K'_j|}])^{1/d}.
\end{align*}
%We briefly mention that as an alternative to using equal weights $ 1/d $, we can apply H\''older's inequality with weights $ p_{1n}, %\dots, p_{dn} $ to arrive at
%\begin{align*}
%\expect{2^{-\sum_{j=1}^d \max\{ K_{j}, K'_j\}}}
%& \leq \prod_{j\in[d]}[\expect{2^{-\frac{1}{p_{j}}\max\{ K_{j}, K'_j\}}}]^{p_{j}} \\
%& \leq 2^{-\lk}\prod_{j\in[d]}[\expect{2^{-\frac{1}{2p_{j}}|K_{j}-K'_j|}}]^{p_{j}}.
%\end{align*}
With $ K_j $ conditionally distributed $ \Binom(\lk, p_{j}) $ given $ \bX $, Biau uses the previous inequality together with the fact that, for $ d \geq 2 $,
\begin{align*}
\Expect_{\Theta, \Theta'}[2^{-\frac{d}{2}|K_{j}-K'_j|}]
& \leq \Prob_{\Theta, \Theta'}(K_j = K'_j) + \Expect_{\Theta, \Theta'}[2^{-|K_1-K'_j|}\indc{K_1 \neq K'_j}] \\
& = \Prob_{\Theta, \Theta'}(K_j = K'_j) + 2\Expect_{\Theta, \Theta'}[2^{-(K_j-K'_j)}\indc{K_j > K'_j}] \\
& \leq 2\Expect_{\Theta, \Theta'}[2^{-(K_{j}-K'_j)}\indc{K_{j} \geq K_{j}}] \\
& \leq \frac{12}{\sqrt{\pi p_{j}(1-p_{j}) \log_2 k_n}},
\end{align*}
where the last inequality follows from \cite[Proposition 13]{biau2012}, to conclude that the estimation error is of order $ O((k_n/n)(\log_2 k_n)^{-d_0/(2d)}) $.
%\footnote{The bound in \cite[Proposition 13]{biau2012} is actually for $ \expect{ 2^{-d(K_{j}-K'_j)_+}} $, which is at least $ \prob{K_{j} \leq K'_j} = \frac{1+\prob{K_{j} = K'_j}}{2} \geq 1/2 $. Nevertheless, an inspection of the proof reveals that the bound therein is valid for $ \expect{2^{-\frac{d}{2}|K_{j}-K'_j|}} $. Incidentally, \cite[Section 3, p. 5]{breiman2004} also appears to have the same oversight.}
Our approach is different. Instead of reducing the calculations so that the expectations involve only their marginals $ K_j $ and $ K'_j $, we work with their joint multinomial distribution.
\end{remark}

\begin{remark}
Compare our result with \cite[Proposition 2]{biau2012}, which shows that the estimation error of $ \widehat Y $ is $ O((k_n/n)(\log k_n)^{-d_0/(2d)}) $. In particular, we improve the exponent in the logarithmic factor from $ d_0/d $ to $ d_0-1 $ (which is a strict improvement whenever $ d_0 > 2 $). In the fully grown case when $ k_n = n $ (i.e., when there is on average one observation per terminal node), the estimation error still decays at a reasonably fast rate $ O((\log n)^{-(d_0-1)/2}) $, due to the correlation between trees.
\end{remark}

\begin{remark}
It is a standard result for partitioning based regression predictors that the estimation error is of order $ \sigma^2 k_n/n $ and hence our improvement \prettyref{eq:variance} is only in terms of the logarithmic factor $ (\log n)^{(d_0-1)/2} $. Note that if the split probabilities $(p_j)_{1 \leq j\leq d}$ are uniform over the $ d $ input features, the logarithmic factor multiplying $ 12\sigma^2 k_n/n $ is small if the tree depth $ \lk $ is greater than a constant multiple of $ d $, i.e., 
$$
\frac{8^d}{\sqrt{(1/d)^d \log^{d-1}_2 k_n}} \ll 1 \qquad \Leftrightarrow \qquad k_n \gg 2^{64 d},
$$
Thus, the improvement manifests for trees with at least $ 2^{64 d} $ terminal nodes. However, with these specifications for $ (p_j)_{1\leq j\leq d} $, the leading term in the approximation error bound \prettyref{eq:bias} is $ S^2L^2 k_n^{-d^{-1}/\log 2} $—which is small precisely when $ k_n \gg 2^{64 d} $—so the improvement to the estimation error is in fact always present in the regime of interest for small mean squared prediction error.
\end{remark}

%\begin{remark}
%\cite[Equation 3]{breiman2004} makes the claim (without proof) that \prettyref{eq:corrbound} is $ O((\log k_n)^{-d_0/2}) $, although in view of the multivariate normal approximation to the multinomial distribution, we see that this cannot be the case since $ \sum_{j=1}^d | K_{j}- K'_j| $ has only $ d_0-1 $ degrees of freedom.
%\end{remark}

\subsection{Median random forests}
Following the same path as the previous subsection, here we derive bounds and rates of convergence for the mean squared prediction error of a median random forest.
\begin{theorem}[Median random forests]\label{thm:medianrisk}
Suppose $ n_0 2^{-\lk} \geq 1 $. Then, under \prettyref{ass:data} and \prettyref{ass:function},
\begin{equation*}
\mathbb{E}[(\widehat Y(\bX)- f(\bX))^2] \leq 256 S^2L^2k^{2\log_2(1-d^{-1}/2)}_n + 2\sigma^2 k_n/n.
\end{equation*}
Consequently, if $ \alpha_d \coloneqq \frac{2\log_2(1-d^{-1}/2)}{2\log_2(1-d^{-1}/2)-1} $ and $ k_n \asymp n^{1-\alpha_d} $, then there exists a constant $ C > 0 $, depending only on B, $ S $, $ d $, $ L $, and $ \sigma^2 $ such that 
\begin{equation*}
\mathbb{E}[(\widehat Y(\bX)- f(\bX))^2] \leq C n^{-\alpha_d}.
\end{equation*}
\end{theorem}

\begin{proof} 
%\paragraph{Approximation error.}

We follow the proof of \cite[Lemma 6.1]{duroux2018impact}, but with some important modifications. Let $\bx \in [0,1]^d$ and let $ \mathcal{C} = \{ N_0, N_1, \hdots, N_{2^{\lk}} \}$ be the number of points in the successive nodes containing $\bx$ (for example, $N_0$ is the number of points in the root node of the tree, i.e., $N_0 = n_0$). We also let $ j_k $ denote the feature index selected at the $ k^{\text{th}} $ step. The counts in $\calC$ implicitly depend on $\mathcal{D}_n$ and $\Theta$, but we suppress these dependencies for clarity. Then $ b_{j}-a_{j} $ can be written as a product of independent beta distributions:  
\begin{align*}
b_{j}-a_{j} \overset{\mathcal{D}}{=} \prod_{k=1}^{\lk} B_k^{\indc{j_k=j}},
\end{align*}
where $ B_k $ are independent $ \text{Beta}(n_k +1,n_{k-1} - n_k) $.
% and the indicator $\indc{j_k=j} $ equals $ 1 $ if the $k^\text{th}$ split of the node containing $\bx$ is performed along the $j^\text{th}$ dimension and equals zero otherwise. 
Consequently, 
\begin{align*}
\Expect_{\Theta|\mathcal{C}}[b_{j}-a_{j}] & = \prod_{k=1}^{\lk} \Expect_{\Theta|\mathcal{C}}\Big[B_k^{ \indc{j_k=j} }\Big] = \\
& = \prod_{k=1}^{\lk} \Big( \frac{d-1}{d} + \frac{1}{d} B_k \Big),
%& = \prod_{k=1}^{\lk} ( \frac{d-1}{d} + \frac{1}{d} \frac{n_k + 1}{n_{k-1} + 1} ) \\
%& \leq \prod_{k=1}^{\lk} ( 1- \frac{1}{d} + \frac{1}{2d}\frac{n_{k-1} + 2}{n_{k-1} + 1} ),
\end{align*}
since $ \Prob_{\Theta|\mathcal{C}}(j_k = j) = 1/d $. Now, by Jensen's inequality for the square function,
$$
\Expect_{\bX_1, \dots, \bX_n}[(\Expect_{\Theta}[b_{j}-a_{j}])^2] \leq \Expect[(\Expect_{\Theta|\mathcal{C}}[b_{j}-a_{j}])^2] = \Expect_{\calC}[\Expect_{\bX_1,\dots, \bX_d | \calC}[(\Expect_{\Theta|\mathcal{C}}[b_{j}-a_{j}])^2]].
$$
Furthermore,
\begin{align*}
& \Expect_{\bX_1, \dots, \bX_n | \mathcal{C}}[(\Expect_{\Theta| \mathcal{C}}[b_{j}-a_{j}])^2] \\ & = 
\prod_{k=1}^{\lk}\Expect_{\bX_1, \dots, \bX_n | \mathcal{C}}\Big[\Big( \frac{d-1}{d} + \frac{1}{d} B_k \Big)^2\Big]  \\ & = \prod_{k=1}^{\lk}\Big(1-\frac{2}{d}+\frac{1}{d^2}+2\Expect_{\bX_1, \dots, \bX_n | \mathcal{C}}[B_k]\Big(\frac{1}{d}-\frac{1}{d^2}\Big) + \Expect_{\bX_1, \dots, \bX_n | \mathcal{C}}[B^2_k]\frac{1}{d^2}\Big).
\end{align*}
We must calculate the first and second moments of a beta distribution in the above expression. Doing so yields
$$
 \prod_{k=1}^{\lk}\Big(1-\frac{2}{d}+\frac{1}{d^2}+2 \frac{n_k + 1}{n_{k-1} + 1}\Big(\frac{1}{d}-\frac{1}{d^2}\Big) + \frac{(n_k + 1)(n_k+2)}{(n_{k-1} + 1)(n_{k-1}+2)}\frac{1}{d^2}\Big).
$$
Next, we use the relation $n_k \leq n_{k-1}/2$ to further bound the above expression by
\begin{align}
& \prod_{k=1}^{\lk}\Big(1-\frac{2}{d}+\frac{1}{d^2}+ \frac{n_{k-1} + 2}{n_{k-1} + 1}\Big(\frac{1}{d}-\frac{1}{d^2}\Big) + \frac{(n_{k-1}+ 2)(n_{k-1}+4)}{(n_{k-1} + 1)(n_{k-1}+2)}\frac{1}{4d^2}\Big) \nonumber \\ & = 
\prod_{k=1}^{\lk}\Big(\Big(1-\frac{1}{2d}\Big)^2 + \frac{4d-1}{4d^2(n_{k-1}+1)}\Big) \nonumber \\
& \leq \prod_{k=1}^{\lk}\Big(\Big(1-\frac{1}{2d}\Big)^2 + \frac{1}{d(n_{k-1}+1)}\Big). \label{eq:final}
\end{align}
Now, $ n_k \geq n_{k-1}/2-1 $ and hence by induction, $ n_k \geq (1/2)^k n_0 -2 $. Furthermore, by assumption $ n_0 2^{-\lk} \geq 1 $, and hence
\begin{equation} \label{eq:combine}
\frac{1}{n_{k-1}+1} \leq \frac{1}{(1/2)^{k-1} n_0-1} \leq \frac{1}{(1/2)^{k-1-\lk}-1} \leq 2^{k-\lk}.
\end{equation}
Continuing from \prettyref{eq:final} and using \prettyref{eq:combine}, we have
\begin{align*}
& \log_2\prod_{k=1}^{\lk}\Big(\Big(1-\frac{1}{2d}\Big)^2 + \frac{1}{d(n_{k-1}+1)}\Big) \\ & \leq \log_2\prod_{k=1}^{\lk}\Big(\Big(1-\frac{1}{2d}\Big)^2 + \frac{2^{k-\lk}}{d}\Big) \\
& = 2\lk\log_2(1-d^{-1}/2) + \sum_{k=1}^{\lk}\log_2\Big(1+\frac{2^{k-\lk}}{d(1-\frac{1}{2d})^2}\Big) \\
& \leq 2\lk\log_2(1-d^{-1}/2) + \frac{2}{d(1-\frac{1}{2d})^2}\\
& \leq 2\lk\log_2(1-d^{-1}/2) + 8.
\end{align*}
This shows that 
$$
\Expect_{\bX_1, \dots, \bX_n}[(\Expect_{\Theta}[b_{j}-a_{j}])^2]  \leq 2^8(1-d^{-1}/2)^{2\lk} \leq 256k_n^{2\log_2(1-d^{-1}/2)},
$$
and hence by \prettyref{thm:approximation error}, the approximation error $ \mathbb{E}[(\overline Y(\bX) - f(\bX))^2] $ is at most
$$
S\sum_{j=1}^d\|\partial_j f\|^2_{\infty}256k_n^{2\log_2(1-d^{-1}/2)} \leq 256S^2L^2k_n^{2\log_2(1-d^{-1}/2)}.
$$
%\paragraph{Estimation error.}
Finally, it is shown in \cite[Section 6.2]{duroux2018impact} that the estimation error has the bound $ \mathbb{E}[(\widehat Y(\bX)-\overline Y(\bX))^2] \leq 2\sigma^2k_n/n $.
\end{proof}

In \prettyref{tab:rates}, we catalogue our improvements in \prettyref{thm:centerrisk} and \prettyref{thm:medianrisk} to \cite{biau2012} and \cite{duroux2018impact} in terms of the estimation, approximation, and prediction errors of a $ k_n $ that optimizes our upper bounds on the tradeoff between the goodness-of-fit and complexity. To make more the comparisons between the two random forest models easier to see, we consider the agnostic choice $ p_j = 1/d $ for all $ j \in \{1, 2, \dots, d \} $ for centered random forests, producing $ p = 1/d $. For the sake of clarity, we also ignore logarithmic factors in $ n $ and replace the rate $ \frac{2\log_2(1-d^{-1}/2)}{2\log_2(1-d^{-1}/2)-1} $ with the more transparent lower bound $ \frac{1}{d\log 2 + 1} $.

\begin{remark} According to \cite[Example 6.5]{yang1999}, the minimax rate for Lipschitz regression models in $ d $ dimensions is $ \Theta(n^{\frac{2}{d+2}}) $. Thus, we see our rate $ \frac{2\log_2(1-d^{-1}/2)}{2\log_2(1-d^{-1}/2)-1} $ for median and centered random forests is minimax optimal \emph{only} when $ d = 1 $.
\end{remark}

\begin{remark}
Compare our choice $ k_n = \Theta(n^{\frac{d\log 2}{d\log 2+1}}) $ with that of \cite[Corollary 6]{biau2012} and \cite[Theorem 3.1]{duroux2018impact}, namely, $ k_n = \Theta(n^{\frac{d(4/3)\log 2}{d(4/3)\log 2+1}}) $. Thus, a better prediction error bound is achieved if the trees are shallower. 
\end{remark}

\begin{table}[h!]
\centering
\begin{tabular}{c | c | c | c | c|}
 & Approximation error & Estimation error &  $ k_n $ & Rate \\ 
 \hline
\cite{biau2012, duroux2018impact} & $ k_n^{-\frac{1}{d(4/3)\log 2}} $ & $ k_n/n $ & $ n^{\frac{d(4/3)\log 2}{d(4/3)\log 2+1}} $ & $ n^{-\frac{1}{d(4/3)\log 2+1}} $  \\ 
Improvement & $ k_n^{-\frac{1}{d\log 2}} $ & $ k_n/n $  & $ n^{\frac{d\log 2}{d\log 2+1}} $ & $ n^{-\frac{1}{d\log 2 + 1}} $ \\
 \hline
\end{tabular}
\caption{Comparison of convergence rates to \cite{biau2012} and \cite{duroux2018impact}.}
\label{tab:rates}
\end{table}

%The leading terms in the risk bound from \prettyref{thm:risk} are $ S^2L^2k^{2\log_2(1-p_n/2)}_n $ and \\ $ \frac{12\sigma^2k_n}{n}\frac{(8S)^{S-1}}{(1+\xi_n)^{S-1}\sqrt{\log^{S-1}_2 k_n}} $. Optimizing their sum over $ k_n $ leads to the following corollary.
%\begin{corollary} \label{cor:main}
%Assume the same setup as \prettyref{thm:risk}. Let $ \alpha_S = \frac{2\log(1-d^{-1}/2)}{2\log(1-d^{-1}/2)-\log 2} $ and $ k_n \asymp (n(\log^{d-1}_2 n)^{1/2})^{1-\alpha_S} $. There exists a constant $ C > 0 $, depending only on B, $ S $, $ L $, and $ \sigma^2 $, such that
%\begin{equation*}
%\expect{\widehat Y(\bX) - f(\bX)}^2 \leq VS^2L^2(S^{-S+3}(L/\sigma)^2n(\sqrt{\log n})^{S-1})^{-\alpha_S}.
%\mathbb{E}[(\widehat Y(\bX)- f(\bX))^2] \leq C(n(\log^{d-1}_2 n)^{1/2})^{-\alpha_S}.
%\end{equation*}
%and
%\begin{equation*}
%\expect{\widehat Y(\bX) - f(\bX)}^2 \leq VS^2L^2(S^{-S+3}(L/\sigma)^2n(\sqrt{\log n})^{S-1})^{-\alpha_S}.
%\mathbb{E}[(\widehat Y(\bX)- f(\bX))^2] \leq Cn^{-\alpha_S}.
%\end{equation*}
%\end{corollary}
\begin{remark}
Since theoretically favorable choices of $ k_n $ depend on unknown quantities, in practice, good values can be chosen using cross-validation.
\end{remark}

%As with \cite[Remark 10]{biau2012}, the assumption of uniform design is not crucial to our analysis. If instead $ \bX $ has density $ f(\cdot) $ which satisfies $ 1/c \leq f(\bx) \leq c $ for some universal constant $ c > 0 $ and for all $ \bx \in [0, 1]^d $, the conclusions of \prettyref{thm:risk}, and \prettyref{cor:main} remain true, with only minor adjustments to the constants. The resulting bounds would depend on $ c $ in a way that could be undesirable if $ 1/c $ is extremely small or $ c $ is extremely large.

\begin{figure}[H]
\centering
  \includegraphics[width=0.5\linewidth]{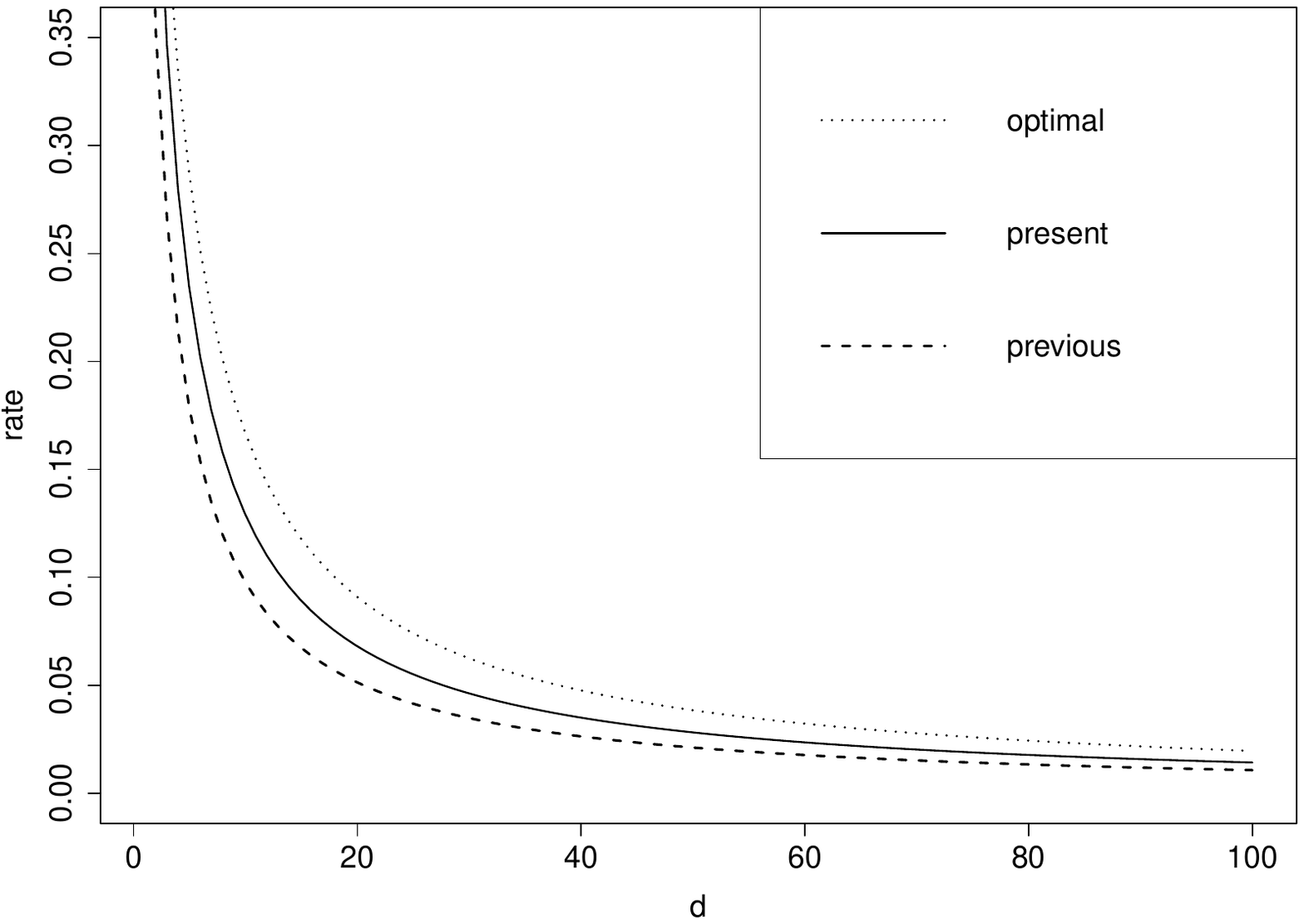}
\caption{A plot of the previous rate $ \frac{1}{d(4/3)\log 2+1} $ from \cite[Corollary 6]{biau2012} and \cite[Theorem 3.1]{duroux2018impact}, the new rate $ \frac{1}{d\log 2 + 1} $ from \prettyref{thm:centerrisk} and \prettyref{thm:medianrisk}, and the minimax optimal rate $ \frac{2}{d+2} $ \cite[Example 6.5]{yang1999}.}
\label{fig:rate}
\end{figure}

\subsection{Data-driven approach for split probabilities} \label{sec:data}

 The approximation error upper bound \prettyref{eq:bias} involves a subtle interplay between the split probabilities $ (p_j)_{1 \leq j \leq d} $ and the size of the partial derivatives of the regression function—directions that have larger variability require more splits—and thus have higher selection probabilities. If each direction contributes equally to the variability of the regression function, then (by a Lagrange multipliers argument)
$$
\sum_{j=1}^d\|\partial_j f \|^2_{\infty}k^{2\log_2(1-p_{j}/2)}_n \approx L^2\sum_{j\in\calS} k^{2\log_2(1-p_{j}/2)}_n
$$
is minimized when the $ (p_j)_{1 \leq j \leq d} $ are uniform over the set of relevant features, i.e., $ p_j = 1/S $ for $ j \in \calS $ and $ p_j = 0 $ otherwise. When this is the case and $ k_n \asymp n^{1-\alpha_S} $, \prettyref{thm:centerrisk} yields the rate \prettyref{eq:rate}, which beats the minimax optimal rate $ \Theta(n^{-\frac{2}{d+2}}) $ \cite[Example 6.5]{yang1999} for Lipschitz regression models in $ d $ dimensions when $ \alpha_S \geq \frac{2}{S+2} $, or roughly when $ S \leq \lfloor 0.72 d \rfloor $ (cf., $ S \leq \lfloor 0.54 d \rfloor $ from \cite[p. 1069]{biau2012}). 

Since the set $ \calS $ is not known a priori, how can one learn these optimal probabilities from the data? To avoid entanglement with the same data used to train the random forest, one solution is to adaptively select candidate strong features using a second sample $ \calD'_n = \{ (\bX'_1, Y'_1), \dots (\bX'_n, Y'_n) \} $, independent of $ \calD_n $ (which can be done, for example, by sample-splitting). Here, candidate strong features are those that maximize the \emph{decrease in variance} that would be obtained if the node $\bt$ is split along the direction $j$ at position $s$, denoted by $ \widehat\Delta(j, s, \bt) $ \cite[Definition 8.13]{breiman1984} and constructed from the second sample $ \calD'_n $.

%$ \widehat\Delta(j, s, \bt) $ \cite[Definition 8.13]{breiman1984}, constructed from the second sample $ \calD'_n $, within a current node $ \bt $ along direction $ j $ at split point $s$.

%\begin{equation} \label{eq:sse}
%\frac{1}{n}\sum_{i=1}^{n} (Y'_i - \overline Y'_{[a_j, b_j]})^2 \indc{\bX^{'(j)}_i \in [a_j, b_j]} - \frac{1}{n}\sum_{i=1}^{n} (Y'_i - \overline Y'_{A^L_j})^2 \indc{\bX^{'(j)}_i \in A^L_j} - \frac{1}{n}\sum_{i=1}^{n} (Y'_i - \overline Y'_{A^R_j})^2 \indc{\bX^{'(j)}_i \in A^R_j},
%\end{equation}

%\begin{equation} \label{eq:sse}
%\widehat\Delta(j, s, \bt) \coloneqq \frac{1}{N}\sum_{\bX'_i \in \bt_L}(Y'_i - \overline Y'_{\bt_L})^2 + \frac{1}{N}\sum_{\bX'_i \in \bt_R}(Y'_i - \overline Y'_{\bt_R})^2,
%\end{equation}
%where $s$ is a proposed split for a variable $\bX^{'(j)}$ that splits $\bt$ into left and right daughter nodes $\bt_L$ and $\bt_R$ depending on whether $ \bX^{'(j)} \leq s $ or $ \bX^{'(j)} > s $ (i.e., $\bt_L = \{\bX'_i \in \bt: \bX^{'(j)}_i \leq s\} $ and $\bt_R = \{\bX'_i \in \bt: \bX^{'(j)}_i > s\} $), $\overline Y'_{\bt_L}$ is the sample mean for $\bt_L$, $ N $ is the sample size of $ \bt $, and $N_L$ is the sample size of $\bt_L$. Similar definitions hold for $\bt_R$. The infinite sample version of $ \Delt\bt(j, s, \bt) $ is
%\begin{align}
%\Delta(j, s, \bt) & \coloneqq \text{VAR}(Y \mid \bX \in \bt, \; \bX^{(j)} \leq s)\prob{\bX^{(j)} \leq s \mid \bX \in \bt} + \nonumber \\ & \qquad \text{VAR}(Y \mid \bX \in \bt, \; \bX^{(j)} > s)\prob{\bX^{(j)} > s \mid \bX \in \bt}. \label{eq:ssepop}
%\end{align}
To ensure that the candidate strong (resp. weak) features have high (resp. low) split probabilities, first randomly select a subset $ \calM $ of $ M $ of the $ d $ features. Then, for each selected feature, calculate the best split $ \hat s_j \in \argmax_{s}\widehat\Delta(j, s, \bt) $ and store the corresponding value $ \widehat\Delta(j, \hat s_j,\bt) $. Finally, select one feature $ \hat\jmath \in \argmax_{j\in\calM}\Delta(j, \hat s_j, \bt) $ to split along and repeat the procedure again in the daughter nodes. This procedure produces split probabilities $(p_j(\bt))_{1 \leq j\leq d}$ that depend on the individual nodes.
%Define $ p_{j} $ as the probability that the $ j^\text{th} $ variable is selected.
%\begin{enumerate}[(i)]
%\item Select at random, with replacement, $ M_n $ candidate features to split on.
%\item For each of the $ M_n $ elected features, calculate the best split, that is, the split which most
%decreases the within-node sum of squares on the second sample $ \calD'_n $.
%\item Select one variable at random among the features which output the best within-node sum of squares decreases, %and cut.
%\end{enumerate}
As is argued in \cite[Section 3]{biau2012}, by considering the average case behavior, this procedure will ideally produce split probabilities $(p_j)_{1 \leq j\leq d}$ that concentrate approximately around $ 1/S $ for $ j \in \calS $ and zero otherwise.
%, viz.,
%\begin{equation*}
%p_{j} \approx \frac{1}{S}(1 -(1-\frac{S}{d})^{M})(1+\xi_{j}) \approx \frac{1}{S}(1+\xi_{j}), \quad j \in \calS,
%\end{equation*}
%and $ p_{j} \approx \xi_{j} $ otherwise, where $ \xi_{j} = O(k_n/n) $. If the sampling is done without replacement, then $ p_{j} = 1/S $ for $ j \in \calS $ and $ p_{j} = 0 $ otherwise provided $ M_n > d-S $.
The reader is encouraged to consult \cite[Section 3]{biau2012} for further details. One natural question to ask is whether strong and weak features can be distinguished from the size of $ \widehat\Delta(j, s, \bt) $ alone.
%This argument justifies assuming henceforth that the $ p_{j} $ admit such a form.
%Therefore, in the version of the centered random forest outlined here, we do not distinguish between \emph{strong} or \emph{weak} variables as before.
Recently, \cite{klusowski2020sparse} showed that maximizing $ \widehat\Delta(j, s, \bt) $ is equivalent to maximizing the Pearson product-moment correlation coefficient between the response data $ Y $ and decision stump $ \widehat Y(j, s) $ along feature $ X_j $ at split $ s $ given $ \bX \in \bt $, namely, $ |\widehat\rho\,(\widehat Y(j,s), Y \mid \bX \in \bt)|  $.\footnote{The decision stump $ \widehat Y(j, s) $ is equal to the sample mean in one of the daughter nodes depending on whether $ X_j \leq s $ or $ X_j > s $.}
Furthermore, for \emph{any} collection of $ M $ monotone functions $ g_j(X_j) $ for $ j \in \calM $, there is an additive model $ Y_0  $ of the form $ \sum_{j\in\calM}\pm g_j(X_j) $ (for example, $ Y_0 $ could be a linear model) such that, almost surely,
$$
|\widehat\rho\,(\widehat Y(\hat\jmath, \hat s_{\hat\jmath}), Y \mid \bX \in \bt)| \geq \frac{1}{\sqrt{M \times (1+\log(2N(\bt)))}} \times |\widehat\rho\,(Y_0, Y \mid \bX \in \bt)|,
$$
where recall that $ N(\bt) $ is the number of data points in the node $ \bt $.
Thus, the size of $ |\widehat\rho\,(\widehat Y(\hat\jmath, \hat s_{\hat\jmath}), Y \mid \bX \in \bt)| $ is approximately $ \Omega(1/\sqrt{M\log N(\bt) }) $ times the correlation between the response data $ Y $ and the additive model $ Y_0 $—which can be made large by appropriate choices of $ g_j(X_j) $. On the other hand, if $ Y $ does not depend locally on $ X_j $, then \cite[Lemma 1]{li2019debiased} show that with high probability,
$$
|\widehat\rho\,(\widehat Y(j, \hat s_j), Y \mid \bX \in \bt)| = O(\sqrt{(\log N(\bt))/N(\bt)}).
$$
Thus, one can distinguish between strong and weak features from the splitting criterion when $ Y $ equals or is strongly (locally) correlated with additive models of the form $ Y_0 = \sum_{j\in\calM}\pm g_j(X_j) $, i.e.,
\begin{equation} \label{eq:cor} |\widehat\rho\,(Y_0, Y \mid \bX \in \bt)| = \Omega(\log N(\bt)\sqrt{\frac{M}{N(\bt)}}). \end{equation} Note that \prettyref{eq:cor} is more likely to occur among shallower nodes when $ N(\bt) $ is large compared to the number of candidate features $ M $.

%\begin{figure} [H]
%\centering
%\begin{subfigure}[t]{0.45\textwidth}
%  \centering
%  \includegraphics[width=1\linewidth]{ratio.pdf}
%\caption{A plot of the ratio $ n^{\frac{d(4/3)\log 2}{d(4/3)\log 2+1}} / n^{1-\alpha_S_d} $ when $ d = 5 $.}
%\label{fig:ratio}
%\end{subfigure}%
%\qquad
%\begin{subfigure}[t]{0.45\textwidth}
%  \centering
%  \includegraphics[width=1\linewidth]{rate.pdf}
%\caption{A plot of the new rate $ \alpha_S $ for $ S = d $ from \prettyref{cor:main} versus the rate $ %\frac{1}{d(4/3)\log 2+1} $ from \cite[Corollary 6]{biau2012} and the minimax optimal rate $ \frac{2}{d+2} %$.}
%\label{fig:rate}
%\end{subfigure}
%\end{figure}

%It will be shown in \prettyref{sec:tight} that the bound for a centered random forest cannot be improved, even under additional smoothness assumptions.

%\begin{remark}
%A more nuanced calculation for centered random forests shows that the approximation error is of order
%$$
%\sum_{j=1}^d \|\partial_j f\|_{\infty}(1-p_{j}/2)^{2\lk},
%$$
%so that the approximation error is \emph{not} minimized when $ p_{j} = 1/S $. That is, $ p_{j} $ should be larger for directions in which the response surface varies more, according to the size of $ \|\partial_j f\|_{\infty} $.
%\end{remark}

\section{Tightness of bounds} \label{sec:tight}

In this section, we show that the approximation error bound \prettyref{eq:bias} for centered random forests we derived in \prettyref{thm:centerrisk} cannot be improved in general. To see this, consider the linear model $Y = \langle \bbeta, \bX \rangle + \varepsilon $,
%\begin{equation*}
%Y = \langle \bbeta, \bX \rangle + \varepsilon,
%\end{equation*}
where $ \bbeta = (\bbeta^{(1)}, \dots, \bbeta^{(d)}) $ is a $ d $-dimensional parameter vector. Then we have the following lower bound on the approximation error of a centered random forest. This lower bound decays with $ k_n $ at the same rate as the estimation error upper bound in \prettyref{thm:centerrisk}, regardless of the split probabilities $(p_j)_{1 \leq j \leq d}$.
%, whose proof we defer to \prettyref{app:appendix}.
\begin{theorem} \label{thm:linear}
%Suppose $ k_n/n \rightarrow 0 $ as $ n \rightarrow \infty $. 
Suppose $Y = \langle \bbeta, \bX \rangle + \varepsilon $, where $ \bbeta = (\bbeta^{(1)}, \dots, \bbeta^{(d)}) $ is a $ d $-dimensional parameter vector. Then, under \prettyref{ass:data} and conditional on $(p_j)_{1 \leq j\leq d}$,
\begin{equation*}
\mathbb{E}[(\overline Y(\bX) - f(\bX))^2] \geq \frac{1}{96}\sum_{j=1}^d|\bbeta^{(j)}|^2 k_n^{2\log_2(1-p_{j}/2)}.
% \frac{S\min_{j \in \calS}|\bbeta^{(j)}|^2}{96}k_n^{2\log_2(1-S^{-1}/2)}.
\end{equation*}
%Furthermore, if $ p_{j} = 1/S $ for $ j \in \calS $ and $ p_{j} = 0 $ otherwise, then
%$$
%\mathbb{E}[(\overline Y(\bX) - f(\bX))^2]\geq \frac{\|\bbeta\|^2_{\calS}}{96}k_n^{2\log_2(1-S^{-1}/2)}.
%$$
\end{theorem}

\begin{proof}
%To alleviate some notational clutter and promote brevity, we will omit dependence of certain quantities on $ \bX $, $ \Theta $, and $ \Theta' $, where $ \Theta' $ is an independent copy of $ \Theta $. Quantities that depend on $ \Theta' $ will be written with a superscript prime in its place. 

%For example, we write $ \bt = \bt $ (resp. $ \bt' = \bt $), $ [a_j, b_j] = [a_j, b_j] $ (resp. $ A'_{j} = [a_j, b_j] $), $ N(\bt) = N(\bt) $ (resp. $ N'_n = N(\bt)(\bX,\Theta') $), $ K_{j} = K_{j} $ (resp. $ K'_{j} = K'_j $), $ \calE = \calE $ (resp. $ \calE'_n = \calE(\bX,\Theta') $), $ B_{j} = B_{j}(\bX) $, $ a_{j} = a_{j} $, and $ b_{j} = b_{j} $. We also define $ \Delta_i = f(\bX_i) - f(\bX) $, for $ i = 1, 2, \dots, n $. \\

Using \prettyref{eq:iden1} from \prettyref{thm:approximation error}, Jensen's inequality for the square function, and exchangeability of the data, we obtain the following lower bound on the approximation error:
\begin{align*}
& \Expect\Big[\Big(\sum_{i=1}^n\Expect_{\Theta}[W_{i}(f(\bX_i) - f(\bX))]\Big)^2\Big] \\ & \geq 
\Expect_{\bX}\Big[\Big(\sum_{i=1}^n\Expect_{\bX_1,\dots,\bX_n,\Theta}[W_{i}(f(\bX_i) - f(\bX))]\Big)^2\Big] \\ & = n^2\Expect_{\bX}[(\Expect_{\bX_1,\dots,\bX_n,\Theta}[W_{1}(f(\bX_1) - f(\bX))])^2].
\end{align*}
%Using an approximation argument similar to \prettyref{eq:generic} reveals that there exist universal constants $ C > 0 $ and $ C' > 0 $ such that
%\begin{align}
%& \expect{\Expect_{\Theta}[W_{1}]\Delta_1 \Expect_{\Theta}[W_{2}]\Delta_2 } \nonumber \\
%& = C(k_n/n)^2\expect{(\expect{\indc{\bX_1 \in \bt}\langle\bbeta, \bX_1 -\bX\rangle \mid \bX})^2 } %\label{eq:leading} \\ & \qquad - C'S\|\bbeta\|^2_{\calS}(k_n/n^3)k^{2\log_2(1-S^{-1}/2)}_n. \nonumber
%\end{align}
Recall the form of the weights
\begin{equation*}
W_{1} = \frac{\indc{\bX_1 \in \bt}}{\sum_{i=1}^n\indc{\bX_i \in \bt}}\Indc_{\calE} = \frac{\indc{\bX_1 \in \bt}}{1+\sum_{i\geq2}\indc{\bX_i \in \bt}}.
\end{equation*}
Define $ T = \sum_{i\geq2}\indc{\bX_i \in \bt} $ and $ \Delta_1 = f(\bX_1) - f(\bX) $.
By a conditioning argument, we write
\begin{align*}
& \Expect_{\bX}[(\Expect_{\bX_1,\dots,\bX_n,\Theta}[W_{1}(f(\bX_1) - f(\bX))])^2] \\
& = \Expect_{\bX}\Big[\Big(\Expect_{\bX_1,\dots,\bX_n, \Theta}\Big[\frac{\indc{\bX_1 \in \bt}\Delta_1}{1+T}\Big]\Big)^2\Big] \\
& = \Expect_{\bX}\Big[\Big(\Expect_{\Theta}\Big[\Expect_{\bX_2, \dots, \bX_n}\Big[\frac{1}{1+T} \Big]\Expect_{\bX_1}\Big[\indc{\bX_1 \in \bt}\Delta_1\Big]\Big]\Big)^2\Big] \\
& = \Expect_{\bX}\Big[\Big(\Expect_{\bX_2, \dots, \bX_n}\Big[\frac{1}{1+T}\Big]\Big)^2\Big(\Expect_{\bX_1, \Theta}[\indc{\bX_1 \in \bt}\Delta_1]\Big)^2\Big],
\end{align*}
where the last line follows from the fact that $ \Expect_{\bX_2, \dots, \bX_n}[\frac{1}{1+T}] $ is independent of $ \Theta $, a consequence of $ T $ being conditionally distributed $ \Binom(n-1, 2^{-\lk}) $ given $ \bX $ and $ \Theta $. Next, we can use Jensen's inequality to lower bound
\begin{equation*}
\Expect_{\bX_2, \dots, \bX_n}\Big[\frac{1}{1+T}\Big] \geq \frac{1}{1+\Expect_{\bX_2, \dots, \bX_n}[T]} = \frac{1}{1+(n-1)2^{-\lk}}.
\end{equation*}
Hence, we obtain that $ n^2\Expect_{\bX}[(\Expect_{\bX_1,\dots,\bX_n, \Theta}[W_{1}(f(\bX_1) - f(\bX))])^2] $ is at least
\begin{equation} \label{eq:quant1}
\Big(\frac{n}{1+(n-1)2^{-\lk}}\Big)^2\Expect_{\bX}[(\Expect_{\bX_1, \Theta}[\indc{\bX_1 \in \bt}\Delta_1])^2].
\end{equation}

Next, in giving a lower bound on $ \Expect_{\bX}[(\Expect_{\bX_1, \Theta}[\indc{\bX_1 \in \bt}\Delta_1])^2] $, we will show that 
\begin{equation} \label{eq:sumuniform}
\Expect_{\bX_1,\Theta}[\indc{\bX_1 \in \bt}\langle\bbeta, \bX_1 -\bX\rangle]
\end{equation}
can be written as a weighted sum of $ d $ independent $ \Unif(0, 1) $ variables minus their mean, $1/2 $. Consequently, the squared expectation of \prettyref{eq:sumuniform} with respect to $ \bX $ is the sum of the respective variances. Using this, we will show that
\begin{equation} \label{eq:num}
\Expect_{\bX}[(\Expect_{\bX_1, \Theta}[\indc{\bX_1 \in \bt}\langle\bbeta, \bX_1 -\bX\rangle])^2] = \frac{2^{-2\lk}\sum_{j=1}^d|\bbeta^{(j)}|(1-p_{j}/2)^{2\lk}}{12}.
\end{equation}
To prove \prettyref{eq:num}, observe that
\begin{align}
& \Expect_{\bX_1}[\indc{\bX_1 \in \bt}\langle\bbeta, \bX_1 -\bX\rangle ] \nonumber \\ & = \sum_{j=1}^d\Expect_{\bX_1}[\indc{\bX_1 \in \bt}(\bbeta^{(j)} (\bX^{(j)}_1 -\bX^{(j)})) ]  \nonumber \\
& = \sum_{j=1}^d\bbeta^{(j)}\prod_{j' \neq j} \lambda([a_{j'}, b_{j'}]) \Expect_{\bX^{(j)}_1}[\indc{\bX_1^{(j)} \in [a_j, b_j]}(\bX^{(j)}_1 -\bX^{(j)}) ]. \label{eq:term}
\end{align}
Next, note that because $ \bX^{(j)} \sim \Unif(0, 1) $, we have
\begin{align*}
& \Expect_{\bX^{(j)}_1}[\indc{\bX_1^{(j)} \in [a_j, b_j]}(\bX^{(j)}_1 -\bX^{(j)})] \\ & = ( b_{j}-a_{j})(\frac{a_{j}+b_{j}}{2} - \bX^{(j)}).
\end{align*}
%where $ a_{j} $ and $ b_{j} $ are the left and right endpoints of $ [a_j, b_j] $. 
Since $ b_{j}-a_{j} = 2^{-K_{j}} $, we have
\begin{align*}
& \Expect_{\bX^{(j)}_1}[\indc{\bX_1^{(j)} \in [a_j, b_j]}(\bX^{(j)}_1 -\bX^{(j)}) ] \\ & = 2^{-K_{j}}\Big(\frac{a_{j}+b_{j}}{2} - \bX^{(j)}\Big).
\end{align*}
Combining this with \prettyref{eq:term} and $ \prod_{j=1}^d 2^{-K_j} = 2^{-\lk} $ yields
\begin{align*}
& \Expect_{\bX_1}[\indc{\bX_1 \in \bt}\langle\bbeta, \bX_1 -\bX\rangle ] \\ & = 2^{-\lk}\sum_{j=1}^d\bbeta^{(j)}\Big(\frac{a_{j}+b_{j}}{2} - \bX^{(j)}\Big).
\end{align*}
Now, by expressions \prettyref{eq:lend} and \prettyref{eq:rend}, which express the endpoints of the interval along the $j^{\text{th}} $ feature as randomly stopped binary expansions of $ \bX^{(j)} $, we have 
\begin{align*}
\frac{a_{j}+b_{j}}{2} - \bX^{(j)} & \eqdistr 2^{-K_{j}-1} - \sum_{k\geq K_{j}+1}B_{k}2^{-k} \\ & \eqdistr 2^{-K_{j}}(1/2 - \sum_{k\geq 1}B_{k+K_{j}}2^{-k}) \\
% \ & \eqdistr 2^{-K_{j}}(1/2 - \sum_{k\geq 1}B_{kj}2^{-k}) \\
& \eqdistr 2^{-K_{j}}(1/2-\tilde{\bX}^{(j)}),
\end{align*}
where $ \tilde{\bX} $ is uniformly distributed on $ [0, 1]^d $. 
Taking expectations with respect to $ \Theta $, we have that
\begin{align}
&\Expect_{\bX_1, \Theta}[\indc{\bX_1 \in \bt}\langle\bbeta, \bX_1 -\bX\rangle ]  \nonumber \\ & \eqdistr 2^{-\lk}\sum_{j=1}^d\bbeta^{(j)}(1-p_{j}/2)^{\lk}(1/2-\tilde{\bX}^{(j)}). \label{eq:sumvar}
\end{align}
Observe that \prettyref{eq:sumvar} is a sum of mean zero independent random variables, and hence, its squared expectation is equal to the sum of the individual variances, viz.,
\begin{align}
& \Expect_{\bX}[(\Expect_{\bX_1, \Theta}[\indc{\bX_1 \in \bt}\langle\bbeta, \bX_1 -\bX\rangle])^2] \nonumber \\ & = 2^{-2\lk}\sum_{j=1}^d|\bbeta^{(j)}|^2(1-p_{j}/2)^{2\lk}\text{VAR}(\tilde{\bX}^{(j)}) \nonumber \\ & = \frac{2^{-2\lk}\sum_{j=1}^d|\bbeta^{(j)}|^2(1-p_{j}/2)^{2\lk}}{12}. \label{eq:quant2}
\end{align}
%Now, if $ p_{j} = 1/S $ for $ j \in \calS $ and $ p_{j} = 0 $ otherwise, then
%$$
%\sum_{j=1}^d|\bbeta^{(j)}|^2(1-p_{j}/2)^{2\lk} = \|\bbeta\|^2_{\calS}(1-S^{-1}/2)^{2\lk}.
%$$
%Note that the optimal value is
%\begin{equation*}
%p_{j} = 1 - \frac{S-1}{|\bbeta^{(j)}|^{\frac{1}{\lk}}\sum_{j=1}^d|\bbeta^{(j)}|^{-\frac{1}{\lk}}},
%\end{equation*}
%which reduces to case $ p_{j} = 1/S $ when all the $ \bbeta^{(j)} $ are equal.
%Putting everything together, it follows that the approximation error is at least a constant multiple of $ \|\bbeta\|^2_{\calS}k^{\log_2(1-S^{-1}/2)}_n $.
Thus, combining \prettyref{eq:quant1} and \prettyref{eq:quant2}, we have shown that
\begin{align*}
\mathbb{E}[(\overline Y(\bX) - f(\bX))^2]  & \geq \Big(\frac{n2^{-\lk}}{1+(n-1)2^{-\lk}}\Big)^2\frac{\sum_{j=1}^d|\bbeta^{(j)}|^2(1-p_{j}/2)^{2\lk}}{12} \\
& \geq \frac{\sum_{j=1}^d|\bbeta^{(j)}|^2k_n^{2\log_2(1-p_{j}/2)}}{96}.
\end{align*}
%To prove the first claim for more general $ p_{j} $, note that 
%$$
%\sum_{j=1}^d|\bbeta^{(j)}|^2(1-p_{j}/2)^{2\lk} \geq \min_{j\in\calS}|\bbeta^{(j)}|^2\sum_{j=1}^d(1-p_{j}/2)^{2\lk}.
%$$
%Using Lagrange multipliers, it can be shown that the convex function $ \sum_{j=1}^d(1-p_{j}/2)^{2\lk} $ is minimized over the probability simplex $ (p_{1},\dots,p_{d}) $ when $ p_{j} = 1/S $ for $ j \in \calS $ and $ p_{j} = 0 $ otherwise. Hence,
%$$
%(\frac{n2^{-\lk}}{1+(n-1)2^{-\lk}})^2(\frac{S}{\sum_{j=1}^d|\bbeta^{(j)}|^{-2/(2\lk-1)}})^{2\lk-1}\frac{S(1-S^{-1}/2)^{2\lk}}{12}.
%$$
\end{proof}

%It should also be mentioned that the assumed form of $ p_{j} = 1/S $ for $ j \in \calS $ in \prettyref{thm:linear} can be relaxed to produce a lower bound of $ \|\bbeta\|^2_{\calS} \,k^{2\log_2(1-\max_{j\in\calS}p_{j}/2)}_n $, provided $ \max_{j\in\calS}p_{j} \geq \max_{j\in \calS^c}p_{j} $.

We also argue that the estimation error bound \prettyref{eq:variance} derived in the proof of \prettyref{thm:centerrisk} is nearly tight when the split probabilities are uniform over all $ d $ features. To this end, in \cite[Theorem 1, Lemma 1, and Theorem 3]{lin2006}, it was shown that if $ w_{max} $ is the maximum number of observations per terminal node, the estimation error for \emph{any} nonadaptive random forest (with uniformly distributed input $ \bX $)\footnote{The lower bound in \cite[Theorem 3]{lin2006} is actually for the mean squared prediction error, but the proof therein is for the variance.} is at least a constant multiple of
\begin{equation} \label{eq:varlower}
\frac{\sigma^2}{w_{max}}\times\frac{(d-1)!}{2^{d}\log^{d-1} n}.
\end{equation}
%By Stirling's formula, \prettyref{eq:varlower} can be further lower bounded by
%\begin{equation} \label{eq:varlowerstirling}
%\frac{C'\sigma^2 w_{max}^{-1}(d-1)^{d-1}}{\log^{d-1} n},
%\end{equation}
%where $ C' = \frac{C\sqrt{2\pi(d-1)}}{(2e)^{d-1}} $. 
Now, the number of observations per terminal node of a centered random forest is on average about $ w_{avg} = n/k_n $ and hence from \prettyref{eq:variance}, centered random forests nearly achieve the best-case estimation error \prettyref{eq:varlower}, namely,
\begin{equation} \label{eq:best}
\frac{\sigma^2}{w_{avg}}\times \sqrt{\frac{(8d)^{d}}{\log^{d-1}(n/w_{avg})}}.
\end{equation}
Taken together, \prettyref{eq:varlower} and \prettyref{eq:best} imply that centered random forests have nearly the lowest estimation error among all purely random forests with nonadaptive splitting schemes. More rigorously, we can prove the following estimation error lower bound, which is valid for any probability sequence $ (p_j)_{1 \leq j \leq d} $.

\begin{theorem}
Suppose $ \lk p_j \geq 1 $. Let $ \calP \coloneqq \{ j : p_j \neq 0 \} $ and $ d_0 \coloneqq \#\calP $. Then, under \prettyref{ass:data} and conditional on $ (p_j)_{1 \leq j \leq d} $,
\begin{equation*}
\mathbb{E}[(\widehat Y(\bX)-\overline Y(\bX))^2] \geq \frac{\sigma^2 k_n}{5n} \frac{(47)^{-d_0}}{\prod_{j\in\calP}p_j \times (\lk)^{d_0-1}}.
\end{equation*}
\end{theorem}
\begin{proof}
First, note that by \cite[Section 5.2, p. 1083-1084]{biau2012},
\begin{align*}
&\mathbb{E}[(\widehat Y(\bX)-\overline Y(\bX))^2] \\ & = n\sigma^2\Expect[(\Expect_{\Theta}[W_{1}])^2] \\
& = n\sigma^2\Expect[\Expect_{\Theta}[W_{1}]\Expect_{\Theta'}[W_{1}]] \\
& = \Expect\Big[\frac{n\sigma^2\indc{\bX_1 \in \bt \cap \bt'}}{(1+\sum_{i=2}^n\indc{\bX_i \in \bt})(1+\sum_{i=2}^n\indc{\bX_i \in \bt'})}\Big] \\
& = \Expect\Big[\frac{n\sigma^2\lambda(\bt \cap \bt')}{(1+\sum_{i=2}^n\indc{\bX_i \in \bt})(1+\sum_{i=2}^n\indc{\bX_i \in \bt'})}\Big],
\end{align*}
where $ \Theta' $ is an independent copy of $ \Theta $.
We first lower bound
\begin{equation*}
\Expect_{\bX_2, \dots, \bX_n}\Big[\frac{1}{(1+\sum_{i=2}^n\indc{\bX_i \in \bt})(1+\sum_{i=2}^n\indc{\bX_i \in \bt'})}\Big].
\end{equation*}
via Jensen's inequality, which yields
\begin{equation*}
\frac{1}{\Expect_{\bX_2, \dots, \bX_n}\Big[\Big(1+\sum_{i=2}^n\indc{\bX_i \in \bt}\Big)\Big(1+\sum_{i=2}^n\indc{\bX_i \in \bt'}\Big)\Big]}.
\end{equation*}
Next, we use linearity of expectation to write
\begin{align*}
&\Expect_{\bX_2, \dots, \bX_n}\Big[\Big(1+\sum_{i=2}^n\indc{\bX_i \in \bt}\Big)\Big(1+\sum_{i=2}^n\indc{\bX_i \in \bt'}\Big)\Big] \\ & = 1 + 2(n-1)2^{-\lk} + (n-1)(n-2)2^{-2\lk} \\  & \qquad + (n-1)\lambda(\bt \cap \bt') \\
& \leq 5n^2/k^2_n,
\end{align*}
where the last inequality follows from $ n \geq 2^{-\lk} $ and $ \lambda(\bt \cap \bt') \leq 2^{-\lk} $. Hence, the estimation error $\mathbb{E}[(\widehat Y(\bX)-\overline Y(\bX))^2]$ can be lower bounded by
\begin{equation} \label{eq:optlower}
\frac{\sigma^2 k_n^2}{5n}\Expect_{\Theta, \Theta'}[\lambda(\bt\cap \bt')],
\end{equation}
where $ \Theta' $ is an independent copy of $ \Theta $. The next key observation is that $ \bt $ and $ \bt' $ are nested according to the maximum of $ K_{j} $ and $ K'_j $, and hence the equality in \prettyref{eq:capiden}. Thus by \prettyref{eq:optlower} and \prettyref{eq:capiden}, we are done if we can show that $ \Expect_{\Theta, \Theta'}[2^{-\frac{1}{2}\sum_{j=1}^d |K_{j} - K'_j| }] $ has a lower bound similar in form to the upper bound in \prettyref{eq:corrbound}. But this follows directly from \prettyref{lmm:multi}, since
\begin{align*}
\Expect_{\Theta, \Theta'}[2^{-\frac{1}{2}\sum_{j=1}^d |K_{j} - K'_j| }] & = \Expect_{\Theta, \Theta'}[2^{-\frac{1}{2}\sum_{j\in\calP} |K_{j} - K'_j|}] \\
& \geq \frac{(47)^{-d_0}}{\prod_{j\in\calP}p_j \times (\lk)^{d_0-1}},
\end{align*}
provided $ \lk p_j \geq 1 $.
\end{proof}

%We do not know of any other random forest model that achieves \prettyref{eq:varlower} or improves upon \prettyref{eq:best}.

Combining the sharpness of our approximation and estimation error bounds for linear models, we conclude that the rate \prettyref{eq:rate} is not generally improvable and hence centered random forests do not achieve the $ d $-dimensional minimax optimal rate $ \Theta(n^{-\frac{2}{d+2}}) $ for $ d $-dimensional Lipschitz regression functions. While centered random forests enjoy near optimal estimation error \prettyref{eq:varlower} (among nonadaptive splitting schemes), their $  O(k^{-\frac{1}{d\log 2}}_n) $ approximation error is far from the optimal $ \Theta(k^{-2/d}_n) $ required to achieve the minimax rate. Only in the one-dimensional setting do centered or median random forests achieve the minimax optimal rate $ \Theta(n^{-2/3}) $ for Lipschitz regression functions in one dimension \cite[Example 6.5]{yang1999}—in the multi-dimensional setting, the rate is suboptimal. These converse statements shed light on the importance of more sophisticated tree construction mechanisms—like Mondrian random forests \cite{mourtada2018}—if optimality is to be guaranteed.

%\section{Supplementary lemmas and their proofs}\label{app:appendix}

%In this supplement, we provide proofs of \prettyref{thm:linear}, \prettyref{lmm:approximation error}, \prettyref{lmm:lowervar}, and \prettyref{lmm:multi}.

%We first mention two useful facts that we will exploit many times:

%\section*{Acknowledgement}

%This research was completed while the author was a visiting graduate student at The Wharton
%School Department of Statistics. The author would like to give special thanks to Mark Low and Tony Cai for providing a space to conduct this work. He is also grateful to Matthew Olson for suggesting relevant literature to review and Edgar Dobriban for helpful discussions.

\appendix
\section{Supplementary results}
\label{app:appendix}

\begin{lemma} \label{lmm:multi}
Let $ \mathbf{M} = (M_1, \dots, M_k) $ be distributed according to a multinomial distribution with $ m $ trials and class probabilities $ (p_1, \dots, p_k) $, each of which is nonzero. Let $ \mathbf{M}' = (M'_1, \dots, M'_k) $ be an independent copy. Then,
\begin{equation} \label{eq:multibound}
\Expect[2^{-\frac{1}{2}\sum_{j=1}^k|M_j-M'_j|}] \leq \frac{8^k}{\sqrt{m^{k-1}p_1\cdots p_k}}.
\end{equation}
%Furthermore, if $ m \geq C'\max_{j\in\calS}p^{-1}_{j} $ for some universal constant $ C' > 0 $, then there exists a universal constant $ C > 0 $ such that
%\begin{equation} \label{eq:multiboundlower}
%\Expect[2^{-\frac{1}{2}\sum_{j=1}^k|M_j-M'_j|}] \geq \frac{C^{k-1}}{\sqrt{m^{k-1}p_1\cdots p_k}}.
%\end{equation}
Furthermore, if $ mp_j \geq 1 $, then
$$
\Expect[2^{-\frac{1}{2}\sum_{j=1}^k|M_j-M'_j|}] \geq \frac{(47)^{-k}}{m^{k-1}p_1\cdots p_k}.
$$
\end{lemma}

\begin{proof}
The proof requires only elementary facts about the multinomial distribution. First, note that
\begin{align}
& \Expect[2^{-\frac{1}{2}\sum_{j=1}^k|M_j-M'_j|}] \nonumber \\ & = \sum_{w_1, \dots, w_{k}}\Prob\Big(\bigcap_{j=1}^{k} \{ |M_j-M'_j| = w_j \}\Big)2^{-\frac{1}{2}\sum_{j=1}^k w_j} \nonumber
\\ & \leq \sum_{w_1, \dots, w_{k-1}}\sum_{\btau \in \{-1, +1\}^{k-1}} \Prob\Big(\bigcap_{j=1}^{k-1} \{ M_j-M'_j = \tau_jw_j \}\Big)2^{-\frac{1}{2}\sum_{j=1}^{k-1} w_j}.
% \nonumber \\
%& = \sum_{\tau \in \{-1, +1\}^{k-1}} \sum_{w_1 \geq 0, \dots, w_{k-1} \geq 0} \Prob(\bigcap_{j=1}^{k-1} \{ \tau_j(M_j-M'_j) = w_j \})2^{-\frac{1}{2}\sum_{j=1}^{k-1} w_j}. 
\label{eq:prob}
\end{align}
Next, let $ p(\bm) = \binom{m}{m_1, \dots, m_k}p_1^{m_1}\cdots p_k^{m_k} $ denote the multinomial mass function and let $ \bm^* $ be ones of its modes.
%where, in each summand of \prettyref{eq:prob}, $ M_k - M'_k = -\sum_{j=1}^{k-1}\tau_j w_j $ and $ w_k = |\sum_{j=1}^{k-1}\tau_j w_j| $.
Then, we can bound each probability in \prettyref{eq:prob} by
$$
 \Prob\Big(\bigcap_{j=1}^{k-1} \{ M_j-M'_j = \tau_jw_j \}\Big) = \sum_{\bm} p(\bm)p(\bm+\btau \bw) \leq p(\bm^*).
$$
Combining these two inequalities, we have
\begin{align}
\Expect[2^{-\frac{1}{2}\sum_{j=1}^k|M_j-M'_j|}] & \leq \sum_{w_1, \dots, w_{k-1}}\sum_{\btau \in \{-1, +1\}^{k-1}}p(\bm^*)2^{-\frac{1}{2}\sum_{j=1}^{k-1} w_j} \nonumber \\ & \leq (4+2\sqrt{2})^{k-1}p(\bm^*). \label{eq:sumineq}
\end{align}
Next, using Stirling's approximation, we have $ m! \leq e \sqrt{2\pi m}(m/e)^m $ and $ m_j! \geq e^{-1}\sqrt{2\pi(m_j+1)}((m_j+1)/e)^{m_j} $. Using these inequalities, we upper bound the multinomial coefficient $ \binom{m}{m_1, \dots, m_k} $, which in turn yields an upper bound on $ p(\bm^*) $, namely,
\begin{equation} \label{eq:pmode}
p(\bm^*) \leq \frac{e^{k+1}}{(\sqrt{2\pi})^{k-1}}\sqrt{\frac{m}{(m^*_1+1)\cdots (m^*_k+1)}}(mp_1/(m^*_1+1))^{m^*_1}\cdots (mp_k/(m^*_k+1))^{m^*_k}.
\end{equation}
Finally, \cite[page 171, Exercise 28, Equation 10.1]{feller1968introduction} states that any mode $ \bm^* $ of the multinomial distribution satisfies $ m^*_j > mp_j -1 $ and hence from \prettyref{eq:pmode},
\begin{equation} \label{eq:final_bound}
p(\bm^*) \leq  \frac{e^{k+1}}{(\sqrt{2\pi})^{k-1}}\frac{1}{\sqrt{m^{k-1}p_1\cdots p_k}}.
\end{equation}
Putting everything together from \prettyref{eq:sumineq} and \prettyref{eq:final_bound}, we have
$$
\Expect[2^{-\frac{1}{2}\sum_{j=1}^k|M_j-M'_j|}] \leq (4+2\sqrt{2})^{k-1}\frac{e^{k+1}}{(\sqrt{2\pi})^{k-1}}\frac{1}{\sqrt{m^{k-1}p_1\cdots p_k}} < \frac{8^{k}}{\sqrt{m^{k-1}p_1\cdots p_k}}.
$$
For the other direction, we first remark that
\begin{equation} \label{eq:explower}
 \Expect[2^{-\frac{1}{2}\sum_{j=1}^k|M_j-M'_j|}] \geq \Prob(\mathbf{M} = \mathbf{M}') = \sum_{\bm} (p(\bm))^2 \geq (p(\bm'))^2,
\end{equation} 
where $ m'_j = \lfloor m p_j \rfloor $.
Following the same strategy as before, we use Stirling's approximation, i.e., $ m! \geq \sqrt{2\pi m}(m/e)^m $ and $ m_j! \leq e\sqrt{2\pi m_j}(m_j/e)^{m_j} $, on the binomial coefficient $ \binom{m}{m_1, \dots, m_k} $, yielding
\begin{align*}
p(\bm') & \geq \frac{e^{-k}}{(\sqrt{2\pi})^{k-1}}\sqrt{\frac{m}{m'_1\cdots m'_k}}(mp_1/m'_1)^{m'_1}\cdots (mp_k/m'_k)^{m'_k} \\ & \geq \frac{e^{-k}}{(\sqrt{2\pi})^{k-1}}\frac{1}{\sqrt{m^{k-1}p_1\cdots p_k}},
\end{align*}
provided $ mp_j \geq 1 $.
%\cite[page 171, Exercise 28, Equation 10.1]{feller1968introduction} also states that $ m^*_j \leq (m+k-1)p_j  $, and hence
%$$
%p(\bm^*) \geq \frac{e^{-k}}{(\sqrt{2\pi})^{k-1}}(1+(k-1)/m)^{-(m+k/2)}\frac{1}{\sqrt{m^{k-1}p_1\cdots p_k}}.
%$$
%If $ m > k $, then $ (1+(k-1)/m)^{-(m+k/2)} \geq (1+(k-1)/m)^{-3m/2} \geq e^{-3k/2} $ and hence
%$$
%p(\bm^*) \geq \frac{e^{-5k/2}}{(\sqrt{2\pi})^{k-1}}\frac{1}{\sqrt{m^{k-1}p_1\cdots p_k}} \geq \frac{(31)^{-k}}{\sqrt{m^{k-1}p_1\cdots p_k}}.
%$$
Applying this inequality to \prettyref{eq:explower} yields
$$
\Expect[2^{-\frac{1}{2}\sum_{j=1}^k|M_j-M'_j|}] \geq \Big(\frac{(e\sqrt{2\pi})^{-k}}{\sqrt{m^{k-1}p_1\cdots p_k}}\Big)^2 \geq  \frac{(47)^{-k}}{m^{k-1}p_1\cdots p_k}.
$$
\end{proof}
%\begin{remark}
%We can also justify \prettyref{eq:multibound} for $ k = 2 $ heuristically using normal approximations. For example, when $ m $ is sufficiently large, $ \frac{M_1-mp_1}{\sqrt{mp_1p_2}} $ is well approximated by a standard normal distribution.  Similarly, $ \frac{M_1 - M'_1}{\sqrt{2mp_1p_2}} $ is well approximated by a standard normal distribution.
%Thus, we have the approximation
%\begin{equation} \label{eq:eq2}
%\Expect[2^{-\frac{1}{2}\sum_{j=1}^2|M_j-M'_j|}] = \Expect[2^{-|M_1-M'_1|}] \approx \Expect[2^{-\sqrt{2mp_1p_2}|Z|}], \qquad Z \sim N(0, 1).
%\end{equation}
%The last expectation in \prettyref{eq:eq2} is equal to the moment generating function $ M_{|Z|}(t) = 2\Prob(Z \leq t)e^{t^2/2} $ of $ |Z| $ evaluated at $ t = -(\log2)\sqrt{2mp_1p_2} $. By the asymptotic expression for Mills's ratio, we have that
%\begin{equation*}
%M_{|Z|}(-t) \sim \frac{2}{\sqrt{2\pi}t}, \quad t \rightarrow \infty.
%\end{equation*}
%Thus, we expect that
%\begin{equation*}
%\Expect[2^{-|M_1-M'_1|}] \approx \frac{1}{\log2\sqrt{\pi mp_1p_2}},
%\end{equation*}
%for large $ m $. Note that this asymptotic expression is within a constant factor of the rigorously established bound in \prettyref{eq:multibound}.
%\end{remark}